\documentclass[pdflatex,sn-mathphys]{sn-jnl}

\jyear{2022}%

\theoremstyle{thmstyleone}%
\theoremstyle{thmstyletwo}%

\theoremstyle{thmstylethree}%

\begin{document}

\title[Skeleton Ground Truth Extraction]{Skeleton Ground Truth Extraction: Methodology, Annotation Tool and Benchmarks}

\author*[1]{\fnm{Cong} \sur{Yang}}\email{cong.yang@suda.edu.cn}
\author[2]{\fnm{Bipin} \sur{Indurkhya}}
\author[3]{\fnm{John} \sur{See}}
\author[4]{\fnm{Bo} \sur{Gao}}
\author[4]{\fnm{Yan} \sur{Ke}}
\author[5,8]{\fnm{Zeyd} \sur{Boukhers}}
\author[6]{\fnm{Zhenyu} \sur{Yang}}
\author[7]{\fnm{Marcin} \sur{Grzegorzek}}



\affil*[1]{\orgname{\small Soochow University}, \city{Suzhou}, \country{China}}
\affil[2]{\orgname{\small Jagiellonian University}, \state{Cracow}, \country{Poland}}
\affil[3]{\orgdiv{\small Heriot-Watt University (Malaysia)}, \state{Putrajaya}, \country{Malaysia}}
\affil[4]{\orgname{\small Clobotics}, \state{Shanghai}, \country{China}}
\affil[5]{\orgname{\small Fraunhofer FIT}, \state{Sankt Augustin}, \country{Germany}}
\affil[6]{\orgname{\small Southeast University}, \state{Nanjing}, \country{China}}
\affil[7]{\orgname{\small University of L\"ubeck}, \state{L\"ubeck}, \country{Germany}}
\affil[8]{\orgname{\small University of Hospital of Cologne}, \state{Cologne}, \country{Germany}}

\abstract{Skeleton Ground Truth (GT) is critical to the success of supervised skeleton extraction methods, especially with the popularity of deep learning techniques. Furthermore, we see skeleton GTs used not only for training skeleton detectors with Convolutional Neural Networks (CNN) but also for evaluating skeleton-related pruning and matching algorithms. However, most existing shape and image datasets suffer from the lack of skeleton GT and inconsistency of GT standards. As a result, it is difficult to evaluate and reproduce CNN-based skeleton detectors and algorithms on a fair basis. In this paper, we present a heuristic strategy for object skeleton GT extraction in binary shapes and natural images. Our strategy is built on an extended theory of diagnosticity hypothesis, which enables encoding human-in-the-loop GT extraction based on clues from the target's context, simplicity, and completeness. Using this strategy, we developed a tool, SkeView, to generate skeleton GT of 17 existing shape and image datasets. The GTs are then structurally evaluated with representative methods to build viable baselines for fair comparisons. Experiments demonstrate that GTs generated by our strategy yield promising quality with respect to standard consistency, and also provide a balance between simplicity and completeness.}

\maketitle

\section{Introduction}
\label{s:intro}
\begin{figure}[t!]
  \centering
    \includegraphics[width=0.8\linewidth]{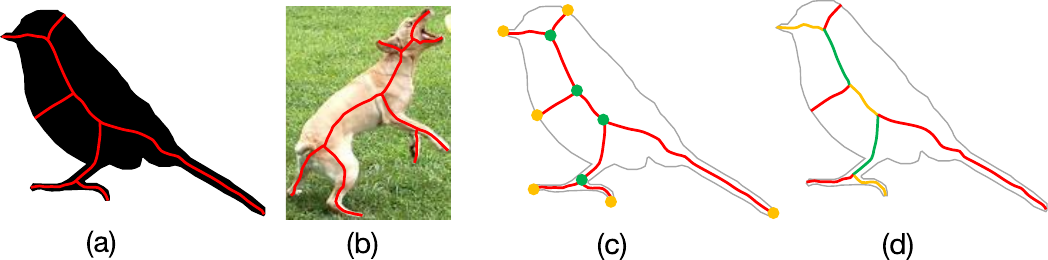}
  \caption{Definition of skeletons and components in a higher level: (a) in binary shape, (b) in natural image, (c) endpoints (orange) and junction points (green), (d) skeleton branches with different colours.}
\label{fig:intro:graphs}
\end{figure}

Skeleton Ground Truth (GT) is critical to the success of supervised skeleton extraction in binary shapes~\cite{Panichev2019} and natural images~\cite{Wang2019DFS} (hereafter referred to as ``shape" and ``image", respectively, see Fig.~\ref{fig:intro:graphs} (a) and (b)). A number of modern skeleton detectors, \emph{i.e.} AdaLSN~\cite{Liu2020ALS} and SkeletonNetV2~\cite{Nathan2021SAD}, are based on Convolutional Neural Networks (CNN), which are trained using skeleton GTs from image and shape datasets, respectively. Moreover, skeleton GT is important to facilitate skeleton-related algorithms such as pruning~\cite{Bai2007SPB}, matching~\cite{Bai2008PSS}, and classification~\cite{Bai2009ICA}. In addition to skeletonization with morphological and geometrical operations~\cite{Giesen2009TSA,Liu2011EGT,Telea2002AAF,Jalba2015AUM,Zhang1984AFP,Ge1996OTG}, skeleton GT extraction should also meet the eye-level view assumption~\cite{Chaz2014PTT} of skeleton simplicity and completeness in different domains. For clarity in terminology, commonly used skeleton components~\cite{Bai2008PSS,Bai2007SPB,Cornea2007CPA} and expressions~\cite{Shen2013SPA} are defined (see Fig.~\ref{fig:intro:graphs} (c) and (d)): 
\begin{itemize}
\item Endpoint: a skeleton point with only one adjacent point.
\item Junction point: a skeleton point with three or more adjacent points.
\item Connection point: a skeleton point that is neither an endpoint nor a junction point.
\item Skeleton branch: a sequence of connection points within two directly connected skeleton points.
\item Skeleton simplicity: higher skeleton simplicity means simpler skeleton structure, e.g. minimal number of branches.
\item Skeleton completeness: higher completeness means a finer-grained representation of object features, e.g. small branches correlated to shape boundary perturbations.
\end{itemize}
\begin{figure}[t!]
  \centering
    \includegraphics[width=0.9\linewidth]{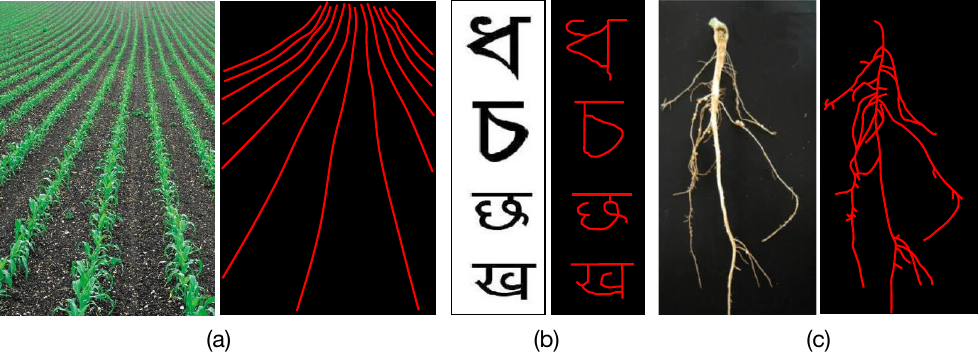}
  \caption{Object skeletons (from simple to complex) in various applications: (a) farmland ridge detection for agricultural robot navigation~\cite{Li2018LSE,Shokouh2021RDB}, (b) character recognition~\cite{Zhang2015STL,Bag2011ROB}, and (3) plant analysis~\cite{Bucksch2014API,Sharma2021PCO}.}
\label{fig:intro:domains}
\end{figure}

As presented in Fig.~\ref{fig:intro:domains}, the requirement of complexity is different in real-world applications~\cite{Saha2016ASO}. For instance, in the scenario of farmland ridge detection for agricultural robot navigation, the ridge skeletons are relatively simple and close to curves). Differently, plant root skeletons are primarily complex, thereby preserving root hair and other details. To properly encode such requirement, skeleton GT extraction is normally addressed by a human-in-the-loop fashion~\cite{Ilke2019SDA}. Particularly, an optimal skeleton GT requires a trade-off between its simplicity and completeness. Thus, following the convention in~\cite{Chaz2014PTT,Lowet2018SSS,Cong2016IOS,Bai2007SPB,Shen2013SPA}, skeleton GT is a satisfaction of the branch simplicity between domain requirements and human perception. An intuitive explanation of such trade-off is that a skeleton GT should satisfy the requirement of simplicity in various domains, while including a proper number of desirable branches (aka. completeness) to preserve object geometrical features. Otherwise, for instance, a skeleton with over-detailed branches could lead to a higher cost on computation and an occurrence of over-fitting problems on matching~\cite{Bai2008PSS}. However, one crucial limitation in existing skeleton GTs lies in the \textit{lack of clarity} and \textit{inconsistency of standards}.

\begin{table}[t!]
\centering
\setlength{\tabcolsep}{4.2pt}
\caption{Comparison of skeleton GT in actively used shape and image datasets. S\&I: Shape and Image. $\surd$ (Yes) and $\times$ (No) denote whether skeleton GT of the full dataset is public available. The size column detail the number of images in each dataset.}
\vspace{-0.9em}
\label{tab:intro:compare}
\renewcommand{\arraystretch}{1.1}
\setlength{\arrayrulewidth}{1pt}
\begin{tabular}{lccclccc}
\hline
Dataset & Type & Size & GT & Dataset & Type  & Size & GT\\ \hline
Animal2000~\cite{Bai2009ICA}  &  Shape & 2000 & $\times$  & ArticulatedShapes~\cite{Haibin2007SCU} & Shape & 40 & $\times$ \\
SkelNetOn~\cite{Ilke2019SDA}  &  Shape & 1725  & $\surd$   & Kimia99~\cite{Sebastian2004ROS} & Shape & 99 & $\times$ \\
Kimia216~\cite{Sebastian2004ROS}  &  Shape & 216  & $\times$ & MPEG7~\cite{Latecki2000SDF} & Shape & 1400 & $\times$ \\
MPEG400~\cite{Cong2014SBO}  &  Shape & 400  & $\times$ & SwedishLeaves~\cite{Soderkvist2001CVC} & Shape & 1125 & $\times$ \\
Tari56~\cite{Asian2005AAR}  &  Shape & 56  & $\times$ & Tetrapod120~\cite{Cong2016OMW} & Shape & 120 & $\times$ \\
SK506~\cite{Shen2016OSE}  &  Image & 506  & $\surd$   & SK1491~\cite{Shen2017DLM} & Image & 1491 & $\surd$ \\
SYMMAX300~\cite{Tsogkas2012LSD}  &  Image & 300 & $\surd$ & SymPASCAL~\cite{Ke2017SSR} & Image & 1435 & $\surd$ \\
EM200~\cite{Yang2014SCO}  &  S\&I & 200  & $\times$   & SmithsonianLeaves~\cite{Haibin2007SCU} & S\&I & 343 & $\times$ \\
WH-SYMMAX~\cite{Shen2016MIS}  &  S\&I & 328  & $\surd$   & \textbf{Our} & \textbf{S\&I} & All & $\surd$ \\ \hline
\end{tabular}
\vspace{-0.6em}
\end{table}
\begin{figure}[t!]
  \centering
    \includegraphics[width=1\linewidth]{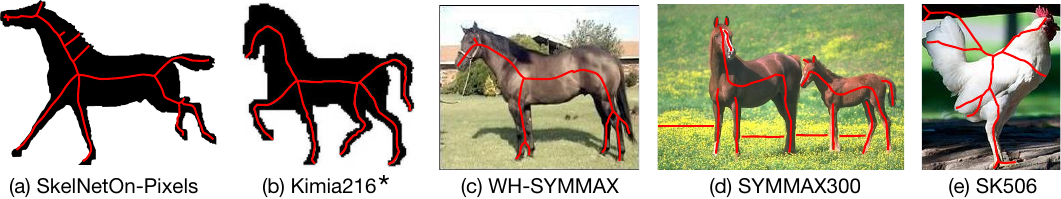}
  \caption{Skeleton GTs in shape (SkeNetOn~\cite{Ilke2019SDA} and Kimia216~\cite{Sebastian2004ROS}) and image (WH-SYMMAX~\cite{Shen2016MIS}, SYMMAX300~\cite{Tsogkas2012LSD} and SK506~\cite{Shen2016OSE}) datasets. \textbf{$\star$}: since there is no public available GT for Kimia216, we take the optimal pruning result of a horse shape presented in~\cite{Bai2007SPB} for comparison.}
\label{fig:intro:gt_comparison}
\vspace{-0.6em}
\end{figure}
\textit{Lack of clarity}: Skeleton GTs of most existing shape datasets are unclear. As presented in Table~\ref{tab:intro:compare}, only two (SkelNetOn~\cite{Ilke2019SDA} and WH-SYMMAX~\cite{Shen2016MIS}) of thirteen actively used shape datasets have publicly available skeleton GTs, though SkelNetOn is only accessible to the registered participates of the SkelNetOn Challenge~\cite{Ilke2019SDA}. For image datasets, skeleton GTs are semi-automatically extracted by object segmentation and skeletonization approaches~\cite{Durix2019TPS,Shen2011SGA}. However, it is unclear whether humans have a similar and stable perception on simplicity and completeness, especially under different contexts from object foreground, background and shape. Context usually refers to the source of contextual associations to be exploited by the visual system~\cite{Oliva2007TRO}. A natural way of representing the context of an object is in terms of its relationship to other objects. In our case, object context is defined as an object's foreground, background, and shape, which are primarily associated with the object skeleton. Theoretically, shape is part of the information in the foreground, while we can easily extract shape by binarizing and filling an object's foreground. Here, we denote shape as an independent context since ten datasets contain only shapes without foreground (see Table~\ref{tab:intro:compare}). In short, there are two uncertainties: (1) it is unclear whether humans have a similar and stable perception of skeleton simplicity and completeness, and (2) it is unclear whether such a perception could be influenced by object foreground and background. Such uncertainties were not structurally studied in existing literature. They can have a tremendous impact on training CNN-based skeleton detectors, making it difficult to compare different skeleton-related algorithms.

\textit{Inconsistency of standards}: We observe glaring inconsistencies among various existing GTs: (1) GT skeletons among existing shape datasets are not always the same. For example, in Fig.~\ref{fig:intro:gt_comparison} (a), the main skeleton branches are shortened whereas some spurious skeleton branches remain in the mouth, neck and hind leg regions. In contrast, in another dataset shown in Fig.~\ref{fig:intro:gt_comparison} (b), only the main branches (not shortened ones) are preserved. (2) GT skeletons from existing image datasets are not consistent. We can clearly see that the GT skeletons in Fig.~\ref{fig:intro:gt_comparison} (d) are in discrete segments, rather than a single connected medial-axis as in Fig.~\ref{fig:intro:gt_comparison} (c). Skeleton GT in Fig.~\ref{fig:intro:gt_comparison} (e) is not accurate. (3) GT skeletons of the shape and the image datasets are not always consistent (Fig.~\ref{fig:intro:gt_comparison} (b) and (c)). Although the main skeleton branches are preserved in both horses, skeleton branches in (c) are shortened. Typically, the shortening of branches may cause blurring between the branches of significant visual parts and branches resulting from noise~\cite{Bai2007SPB}. To sum up, the standards on GT structure (simplicity, completeness, connectivities to branch and boundary) are not consistent. As a result, evaluating skeleton-related pruning, matching and classification approaches with inconsistent GT is an ill-posed problem.

In this paper, we introduce an annotation tool, SkeView, for skeleton GT extraction in image and shape datasets. To do so, we first report an empirical study of human perception on skeleton structure based on the theory of diagnosticity hypothesis~\cite{Tversky1977FOS}. Diagnosticity hypothesis aims to capture the effect of context on target similarity from the perspective of human perception. In our case, exploring human perception on skeleton structure by varying the object context (foreground, background, and shape), time, and participants. Based on these studies, we introduce a general strategy for extracting skeleton GT in image and shape datasets. Our proposed strategy is able to encode human-in-the-loop GT extraction based on clues from the target context, simplicity and completeness. Using this strategy, SkeView is designed and developed to generate skeleton GTs for existing datasets including those in Table~\ref{tab:intro:compare}. Our generated GTs have consistent standards, and properly represent the object geometrical and topological features. These aspects provide a reliable benchmark for assessment. Thus, we can systematically evaluate representative methods using our GTs on skeleton detectors and skeleton-based algorithms, and generate viable baselines for the community.

It should be emphasized that introducing a new skeletonization method is not the focus of this paper, though SkeView can be extended for this purpose. This is because our proposed strategy is applied semi-automatically, and therefore is not suitable for real-time (or quasi real-time) skeleton extraction in various applications. Moreover, desirable properties of skeletons have been well-defined (in 2D at least) via Blum Transform~\cite{Blum1967ATF}, discontinuities of the Distance Transform~\cite{Ge1996OTG}, and many other equivalent definitions from Ogniewicz~\cite{Ogniewicz1992VST}, Telea~\cite{Telea2002AAF}, Latecki~\cite{Latecki2000SDF}, Bai~\cite{Bai2007SPB} and Cornea~\cite{Cornea2007CPA}, etc. Therefore, in this paper, we underscore the suitability of SkeView for training and testing data extraction of skeleton GTs, especially in this era of deep learning. Moreover, skeletons and GTs can be defined in general on a higher level, while it is not possible to find a general definition on a lower level, particularly towards various applications. This is because different applications may have different requirements on skeleton properties (e.g., 2D, 3D, and simplicity). Thus, we also underscore the generalization of our GTs (see Section~\ref{s:tool:gt}) to specific vision tasks in the original datasets, such as skeleton detector training, skeleton matching and shape retrieval, etc.

Succinctly, the main contribution is that we introduce a general strategy to extract skeleton GTs in shape and image datasets. Our strategy meaningfully considers human perception on skeleton simplicity and completeness to adopt various requirements for real-world applications. We present a tool, SkeView, which utilises the proposed methodology to generate skeleton GTs in image and shape datasets. This contributes towards facilitating practical applications and proper benchmarking in future. We also generate skeleton GTs for 17 actively used datasets in Table~\ref{tab:intro:compare} to build new baselines on a consistent and standardized manner. Our comprehensive evaluation demonstrates the efficacy of SkeView, highlighting the need for a new perspective for CNN-based skeleton detectors to become practically relevant and feasible.
\section{Related Works}
\label{s:related}
We present here a brief overview of several existing methods that were proposed for extracting skeleton GTs. For a more thorough treatment on skeletonization methods, compilations by Saha~\cite{Saha2016ASO}, Telea~\cite{Tagliasacchi2016SAS} and Liu~\cite{Liu2011EGT} offer sufficiently good reviews.

\subsection{GT in Shape Datasets}
\label{s:related:shapes}
\begin{figure}[t!]
  \centering
    \includegraphics[width=0.9\linewidth]{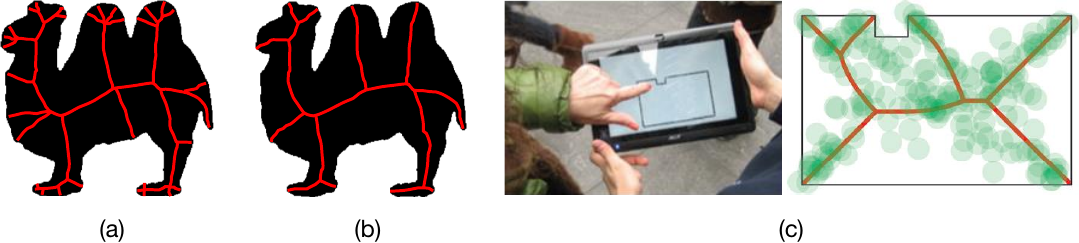}
  \caption{Skeleton GT extraction in shape dataset: (a) automatically with a fixed pruning power~\cite{Bai2007SPB}. (b) semi-automatically with a manual optimized pruning power in SkeView. (c) Purely manual via shape tapping~\cite{Chaz2014PTT}.}
\label{fig:related:generate}
\vspace{-0.7em}
\end{figure}
Fig.~\ref{fig:related:generate} presents existing approaches that could be applied for skeleton GT extraction in shape datasets. As mentioned in Section~\ref{s:intro}, these methods are normally applied semi-automatically to meet human perception on complexity. Otherwise, the extracted GTs are  too simple or contain redundant small branches. For instance, \cite{Bai2007SPB} requires a stop parameter $k$ to control the simplicity of skeleton structures. If $k$ is fixed without manual calibration, redundant small branches are not removed completely in simple shapes, \textit{e.g.} the GT in Fig.~\ref{fig:related:generate} (a) with a fixed $k=30$. In contrast, the GT in Fig.~\ref{fig:related:generate} (b) is extracted based on an optimized $k$ in SkeView. We can clearly see that it is more perception friendly in terms of balancing the skeleton simplicity and completeness.

In contrast to semi-automatic approaches, purely manual GT extraction is conducted with more user interaction, typically using a variety of tools. As shown in Fig.~\ref{fig:related:generate} (c), Firestone {\it et al.}~\cite{Chaz2014PTT} developed an application for a touch-sensitive tablet computer to display single closed geometric shapes, thereby collecting touch data from the participants. Each participate could tap on the displayed shapes anywhere they wished. The collection of their tapped locations provide a global representation of the crowd-sourced perception of major skeletons (aka. GTs). Instead of generating skeletons from scratch, Yang {\it et al.}~\cite{Cong2016IOS} generated a set of GT candidates with different complexity, and then applied a voting scheme based on questionnaires. Each participant was provided with three candidates in a questionnaire, and was asked to select the most promising one, or to draw a new one. Though both these manual approaches can capture crowd-sourced perceptions on skeleton complexity in a proper manner, they are not efficient enough for datasets with a massive number of shapes. Unlike the purely manual approaches, our proposed strategy is more efficient as it generates GT via SkeView semi-automatically and in parallel.

\subsection{GT in Image Datasets}
\label{s:related:images}
In practice, GTs in image datasets are extracted semi-automatically via two steps: segmentation and skeletonisation. The segmentation step is mostly applied manually. For instance, in the SYMMAX300~\cite{Tsogkas2012LSD} dataset, each image was accompanied by 5-7 human segmentations. Thus, multiple binary objects can be obtained for the followed skeletonisation and integration. Although purely manual segmentation can properly ensure the integrity of objects while reducing boundary noises, it is not efficient enough to be applied in practice, particularly preparing massive skeleton GTs for training scenarios. In terms of the skeletonisation step, some existing shape skeleton extraction approaches~\cite{Bai2007SPB,Telea2002AAF,Shen2011SGA} are applied semi-automatically on the shape of segmented objects. As shown in Fig.~\ref{fig:intro:gt_comparison}, these skeleton extraction approaches have different preferences on skeleton geometry and topology. Moreover, it is not clear whether humans have a similar and stable perception of skeleton complexity under different contexts. As a result, skeleton GTs in the existing image datasets are not very consistent (see Fig.~\ref{fig:intro:gt_comparison} (c) (d) (e)). 

In contrast, our proposed method is better in terms of efficiency and consistency. Specifically, our strategy is more general and standardized, as it is built on a structural study of human perception on skeleton GT. Besides, SkeView has an easy-to-use user interface, and a set of convenient functions to improve the efficiency of GT extraction in both shapes and images.
\section{Methodology}
\label{s:med}
Here, we first present a study of human perception of skeleton structure based on the theory of diagnosticity hypothesis~\cite{Tversky1977FOS}. Based on these observations, we introduce a strategy for skeleton GT extraction in the shape and the image datasets.

\subsection{Diagnosticity Hypothesis}
\label{s:com:dia}
\begin{figure}[t!]
  \centering
    \includegraphics[width=0.5\linewidth]{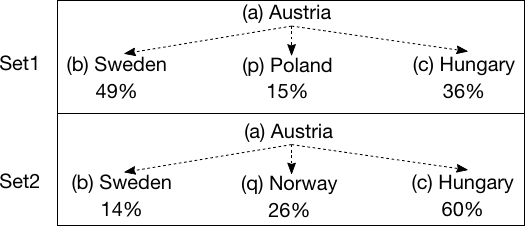}
  \caption{An example of diagnosticity hypothesis~\cite{Tversky1977FOS}. The percentage of participants who selected each country (as most similar to Austria) is presented below the name.}
    \vspace{-0.7em}
\label{fig:com:diagnosticity}
\end{figure}

The diagnosticity hypothesis is a classic framework to explore the relation between similarity and context (or grouping) in the domain of cognitive science~\cite{Skov1986IPD}. Specifically, the diagnosticity hypothesis implies that the change in context, induced by the substitution of an odd element, will change the similarities in a predictable manner. An example is shown in Fig.~\ref{fig:com:diagnosticity}: consider two sets of four countries, which differ in only one of their elements (p and q). The four countries of each set were presented to participants, who were instructed to select the country most similar to Austria (a). Note that this experiment was done in the 1970s, so one has to remember the political map of Europe at that time. The final statistical results are shown in percentages. It is interesting to observe that the selection results in Set 1 and Set 2 are different (Austria (a) is grouped with Sweden (b) in Set 1, and with Hungary (c) in Set 2) by changing only one element (p to q), though both (p) and (q) are not the final results. The diagnosticity hypothesis example in Fig.~\ref{fig:com:diagnosticity} demonstrates that human perception of selection (a country most similar to Austria) could be influenced by a change of context (from Poland to Norway). In our case, human perception of selection (a branch to prune) could be affected by the shift in object contexts, such as shape, foreground, and background.
\begin{figure}[t!]
  \centering
    \includegraphics[width=0.8\linewidth]{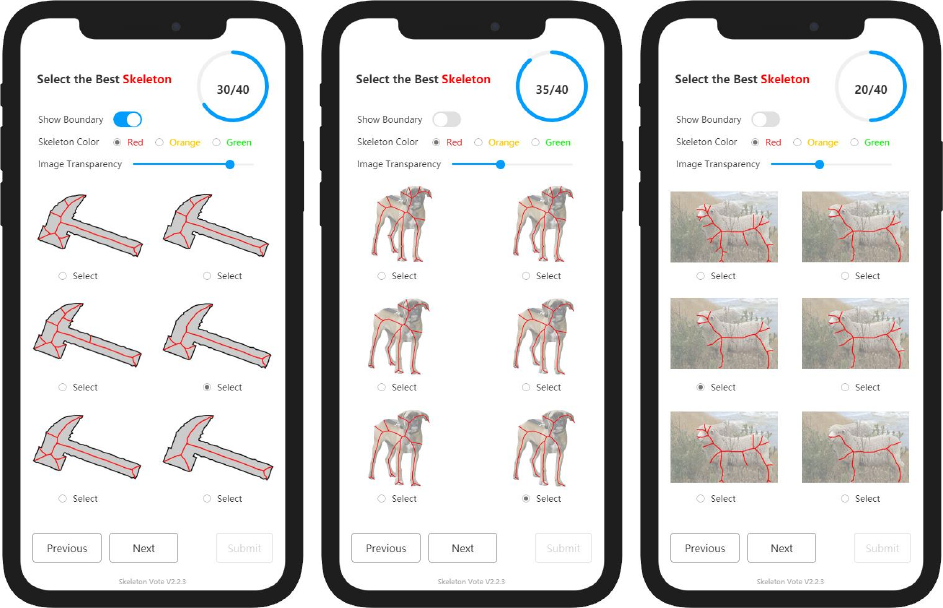}
  \caption{Interfaces of our APP for skeleton selection (best viewed in color).}
    \vspace{-0.8em}
\label{fig:com:app}
\end{figure}

Accordingly, our study was conducted by evaluating the robustness of human perception on skeletons spatially and temporally. In other words, (1) perception of an object skeleton in the context of object shape, segmented foreground and full image, (2) perception of an object skeleton in different time slots, and (3) perception of an object skeleton by different volunteers. Thus, our study is an extension of diagnosticity hypothesis: verifying whether a skeleton GT could be robust for different people, at different times, and in different contexts. Due to the limitations of face-to-face surveying during the global pandemic~\cite{Fanelli2020AAF}, we developed a phone application (APP) to collect perceptions from different participants, as presented in Fig.~\ref{fig:com:app}. Our APP contained four major components: a counter showing processed/remaining images (top right), a setting panel for the boundary, colour and transparency (top left), a selection area for the skeletons (middle), and buttons for page navigation and submission (bottom). In total, 90 volunteers (45 females, 45 males) participated in the study (January to March, 2021), most of whom were students and teachers from Northeast Normal University (NENU), China.

\begin{table}[t!]
\centering
\setlength{\tabcolsep}{4pt}
\caption{Statistics of total endpoint numbers from the most voted skeletons.}
\label{tab:com:study}
\renewcommand{\arraystretch}{1.1}
\setlength{\arrayrulewidth}{1pt}
\begin{tabular}{lccclccclccc}
\hline
 & \multicolumn{3}{c}{Group1} &  & \multicolumn{3}{c}{Group2} &  & \multicolumn{3}{c}{Group3} \\ \hline
Date & Shape & Object & Image & ~ & Shape & Object & Image & ~ & Shape & Object & Image \\
Jan 21 & 378 & - & - &  & - & 330 & - &  & - & - & 315 \\
Feb 04 & - & 326 & - &  & - & - & 320 &  & 380 & - & - \\
Feb 18 & - & - & 314 &  & 373 & - & - &  & - & 322 & - \\
Mar 04 & 378 & - & - &  & - & 330 & - &  & - & - & 315 \\ \hline
\end{tabular}
\vspace{-0.6em}
\end{table}
We randomly selected 30 images from the existing datasets in Table~\ref{tab:intro:compare}, and applied manual segmentation and semi-automatic skeletonization with the method introduced in~\cite{Bai2007SPB}. For some images with complex backgrounds, we intentionally generate two segmented samples, a promising one and a noisy one, for comparison. We generated six skeleton candidates for each shape with different levels of complexity, resulting in a total of $40\times6$ skeletons. To reduce the influence of context from different formats, we organized our volunteers into three groups (30 in each group) and presented object shapes, foregrounds and full images to each group independently. To facilitate the study (see Table~\ref{tab:com:study}), we repeated the survey every two weeks so that the effect of context memorisation could be reduced. For each trial, the skeleton format was changed in each group so that the three formats could be fully surveyed from all groups. We also conducted an additional survey seven weeks later, using the format of the first survey, to measure the stability of results with respect to time passing.

\begin{figure}[t!]
  \centering
    \includegraphics[width=0.8\linewidth]{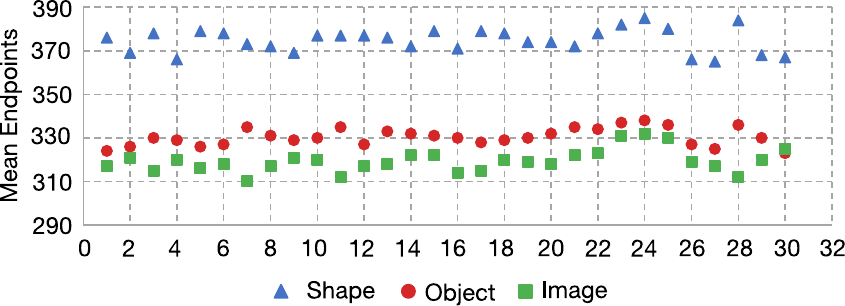}
  \caption{Comparison of each participant in Group2. The participant IDs are shown on the horizontal axis (1 to 30) (best viewed in color).}
\label{fig:com:group2}
\vspace{-0.8em}
\end{figure}
For quantitative analysis, the number of endpoints (more branches implies more endpoints) is used in our study. It should be noted that skeleton simplicity (see Eq.~\ref{exp:simplicity} in Section~\ref{s:tool:properties}) can also be used for the quantitative analysis. Particularly, it has higher discriminative power than the number of endpoints. Here, we employed the number of endpoints in Table~\ref{tab:com:study} for two reasons: (1) The differences between manually voted skeletons from Shape, Object, and Image are distinct, e.g., 378, 326, and 314, respectively. Thus, endpoint statistics are already enough to tell the difference at the coarse-grained level. (2) It is easier to count and visually recheck, particularly in our user study scenario using the questionnaire in APP. Based on the statistics shown in Table~\ref{tab:com:study}, we found that the number of endpoints in shapes, foregrounds and full images (``shape", ``object", ``image") are within $[373, 380]$, $[322, 330]$ and $[314, 320]$, respectively. In other words, each group has a rather consistent perception on skeleton structure, with differences of only about $2\%$. However, as shown in Fig.~\ref{fig:com:group2}, individual perception are varied, ranging from 365 to 385 for shapes, 323 to 338 for objects and 310 to 332 for images. For instance, ID 27 prefers concise skeletons while the perception of IDs 11 and 28 are erratic. We believe the idea of group integration~\cite{Tsogkas2012LSD,Cong2016IOS} produces a more consistent performance than the individual scheme in~\cite{Ke2017SSR,Shen2016MIS,Shen2016OSE,Shen2017DLM}. As the endpoint numbers on January 21 and March 04 were almost the same, we can assume that the human perception of skeleton structure is stable over time. Considering the mean values of shape (377) vs. object (326), we find that the foreground context has a considerable influence on human perception, with about 13.5\% reduction from shape to object formats. However, the difference between object (326) and image (316) is less obvious, with only about 3.1\% reduction.

\begin{figure}[t!]
  \centering
    \includegraphics[width=1\linewidth]{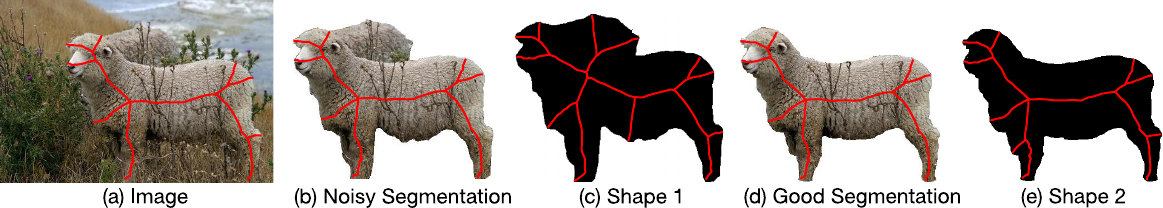}
  \caption{Most selected skeletons of a sheep in the full image, two segmentations, and the correlated shapes, using our APP (best viewed in color).}
\label{fig:com:ske_compare}
\vspace{-0.8em}
\end{figure}
To better understand these results, the most voted skeletons of a sheep image are presented in Fig.~\ref{fig:com:ske_compare} (a), together with its two segmentations (noisy (b) and good (d)) and their corresponding shapes. We intentionally eliminated the fore- and background of the object in (c) and (e) to reduce their context influence. The only difference is that Shape 1 contains noises in the top-left region (head and neck). We find that skeletons in (a), (b) and (d) are almost the same. This is understandable as illusions from the background and the boundary noise can be easily filtered by human inspection. However, as presented in (c) and (e), most volunteers tended to use more skeleton branches to fill their perceptual gaps on shapes (where there is less context information). In cognitive science, the perceptual gap~\cite{Teichmann2021RVM} refers to cognitive biases from information gaps, such as occlusion (internal and external) and misunderstanding, etc. In our case, a perceptual gap occurs since it is difficult to identify the original object (a sheep or something else) from the noisy shape in (c). As a result, volunteers tend to use more skeleton branches to fill their perceptual gaps in this shape. For instance, it is difficult to identify the original object of Shape 1 in Fig.~\ref{fig:com:ske_compare} (c), particularly at the head and neck regions. As a result, the skeleton in Shape (c) is erroneously more extensive than the ones in (b), (d) and (e). Overall, our observations can be summarized as follows:
\begin{itemize}
  \item O1: Perception is robust to the time and volunteer groups.
  \item O2: Perception is robust to segmented objects and images.
  \item O3: Perception of shapes is not robust and is easily influenced by deformations from noises and occlusion.
  \item O4: People tend to use more skeleton branches when there exist perceptual gaps on shapes, and vice versa.
\end{itemize}

These four observations are used to design the strategy and Graphical User Interface (GUI) of the annotation tool for extracting the skeleton GT in the image and the shape datasets.

\subsection{Strategy}
\label{s:com:strategy}
Given an image $\textbf{I}$, let $\textbf{M}$ and $\widehat{\textbf{M}}$ denote a segmented object and its shape, respectively. Let the final GT skeleton be $\textbf{S}$. In brief, our GT extraction strategy is composed by two steps: preprocessing and pruning. The preprocessing step includes target object segmentation (for image datasets) and initial GT extraction in a coarse level. Then, a heuristic pruning process is conducted semi-automatically based on the above observations (O1-O4) and the human perception on simplicity and completeness.

\begin{figure}[t!]
  \centering
    \includegraphics[width=1\linewidth]{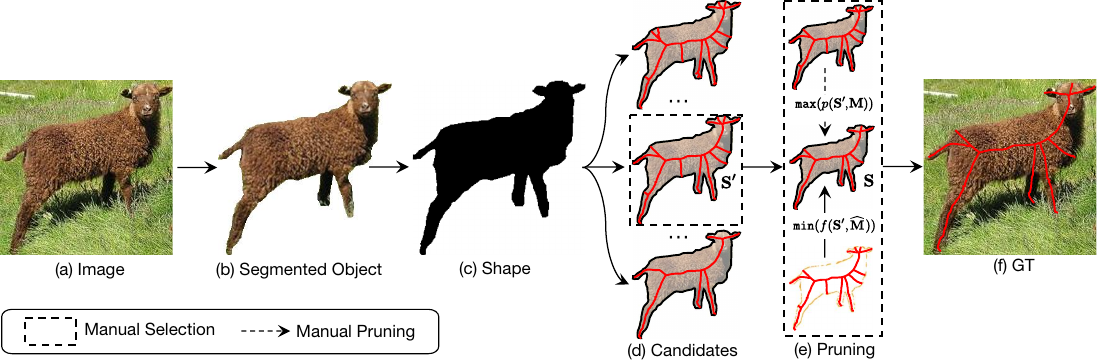}
  \caption{Pipeline of our proposed skeleton GT generation strategy in the image scenario. $p(\cdot)$ and $f(\cdot)$ denote the satisfaction of human perception and shape reconstruction error, respectively (best viewed in color).}
\label{fig:com:image_scenario}
\vspace{-0.7em}
\end{figure}
Such coarse-to-fine strategy can inherently improve the efficiency of GT extraction, as most time-consuming operations are automatically applied in the first step. Specifically, based on the segmented $\widehat{\textbf{M}}$ (Fig.~\ref{fig:com:image_scenario} (b) and (c)) with~\cite{He2017MRC}, the skeletonization approach~\cite{Shen2013SPA} is employed for extracting the initial skeleton. This process effectively reduces the workload of the manual pruning that follows, as most of the redundant branches are removed in the initial skeleton. To bring more flexibility, we intentionally preserve more branches than the optimal ones from the automatic approach to generate a set of candidates with different levels of complexity (Fig.~\ref{fig:com:image_scenario} (d)). $\textbf{S}^{\prime}$ denotes the selected initial skeleton from the candidates.

The second step of skeleton pruning is a heuristic and a semi-automatic process to identify the skeleton GT: that is, maximizing the (human) perceptual simplicity while keeping the skeleton as much complete as possible. For simplicity, Fig.~\ref{fig:com:image_scenario} (d) and (e) depicts how the skeletons appear during the selection and the pruning processes. This is motivated by our observations in O2 and O3. For completeness, inspired by~\cite{Shen2013SPA}, we introduce a shape reconstruction error to represent the skeleton completeness: that is, keeping the reconstruction error of $\textbf{S}$ to $\widehat{\textbf{M}}$ as small as possible (Fig.~\ref{fig:com:image_scenario} (e)). Then, the skeleton GT $\textbf{S}$ is extracted by:
\begin{equation}
\small
\textbf{S}_{image} = \mathtt{max}(p(\mathtt{min}(f(\textbf{S}^{\prime},\widehat{\textbf{M}})), \textbf{M}))
\quad .
\label{equ:per:image}
\end{equation}
where $p(\cdot)$ and $f(\cdot)$, respectively, represent the (human) perceptual satisfaction and the shape reconstruction error. Thus, Eq.~\ref{equ:per:image} is a semi-automatic annotation method as the variable $f(\cdot)$ is computed automatically and $p(\cdot)$ is determined by a human during manual pruning (\textit{i.e.} manual selection of branch candidates for pruning). That is, calculating $f(\cdot)$ to inspire a human on trading-off the skeleton simplicity (domain requirement) and completeness ($f(\cdot)$ value). As a result, $p(\cdot)$ and $f(\cdot)$, respectively, are inherently maximized and minimized.

The rational behind Eq.~\ref{equ:per:image} is that, as O4 suggests, people intend to use fewer branches (simple skeleton) on $\textbf{I}$ and $\textbf{M}$. This applies the diagnosticity hypothesis, whereby factors from other contexts ({\it i.e.} the reconstruction error) could potentially influence human perception. In practice, $p(\cdot)$ is maximized by dynamically selecting and pruning branches based on the eye-level view assumption of skeleton simplicity, and hints from the reconstruction error $f(\textbf{S}^{\prime},\widehat{\textbf{M}})$:
\begin{equation}
\small
f(\textbf{S}^{\prime},\widehat{\textbf{M}}) = \frac{\vert\Lambda(\widehat{\textbf{M}}) - \Lambda(R(\textbf{S}^{\prime}))\vert}{\Lambda(\widehat{\textbf{M}})}
\quad .
\label{equ:per:error}
\end{equation}
where $\Lambda(\cdot)$ denotes the area in terms of pixels, $R(\textbf{S}^{\prime})$ is the shape reconstructed from $\textbf{S}^{\prime}$:
\begin{equation}
\small
R(\textbf{S}^{\prime}) = \bigcup_{s\in\textbf{S}^{\prime}}B(s, r(s))
\quad .
\end{equation}
where $r(s)$ is the radius of the maximal disc $B(s, r(s))$ centered at a point $s\in\textbf{S}^{\prime}$. In practice, $r(s)$ is approximated with the
values of the distance transform at $s$. Motivated by the observation in O1, we suggest to conduct observations according to Eq.~\ref{equ:per:image} by at least three participants, and heuristically take $\textbf{S}_{image}$ to be the one with the maximum votes (when 2 skeletons are the same) or median reconstruction error (when 3 skeletons are different). To promote the efficiency of the human-in-the-loop approach, we introduce a new tool, SkeView, in Section~\ref{s:tool} with various functions for segmentation, initialization, pruning, and integration. 
\begin{figure}[t!]
  \centering
    \includegraphics[width=0.68\linewidth]{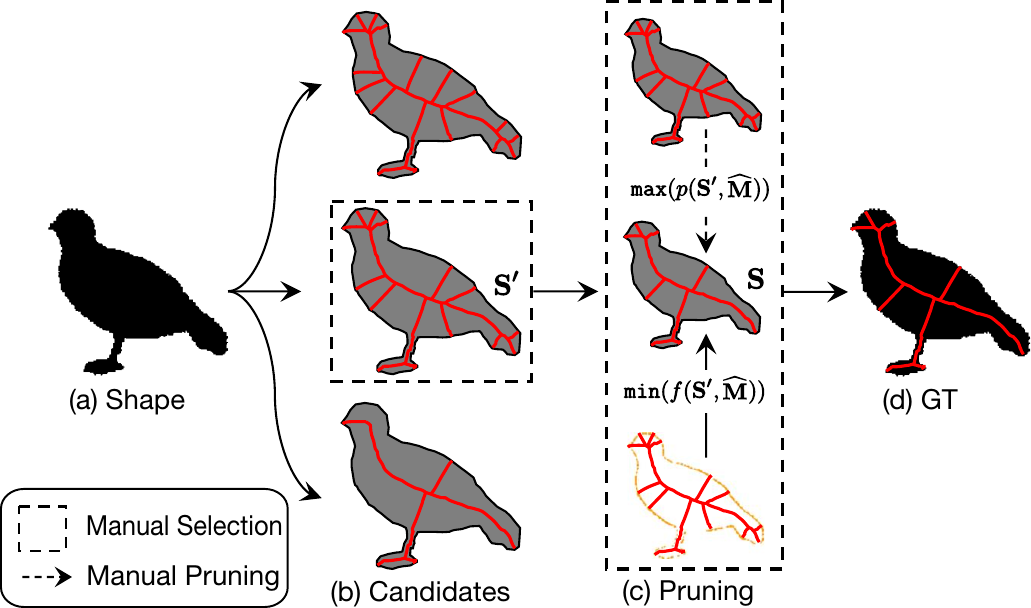}
  \caption{Pipeline of our proposed skeleton GT generation strategy in the image scenario. Best viewed in color.}
\label{fig:com:shape_scenario}
\vspace{-0.7em}
\end{figure}

\begin{figure}[t!]
  \centering
    \includegraphics[width=1\linewidth]{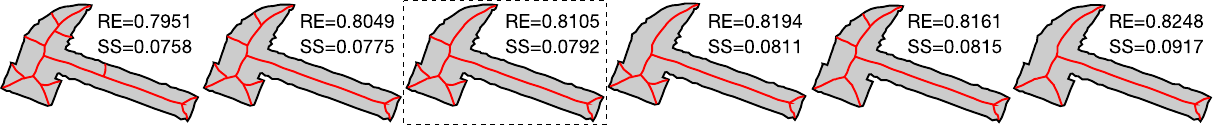}
  \caption{The changes in simplicity (SS) and reconstruction error (RE) during the pruning.}
\label{fig:com:trade_off}
\vspace{-0.8em}
\end{figure}

For the shape scenario, as presented in Fig.~\ref{fig:com:shape_scenario}, our strategy is similar to the workflow from Fig.~\ref{fig:com:image_scenario} (c) to (f). As there is no $\textbf{M}$ displayed below the skeletons, only shape contour and the skeleton are fused in the illustration in Fig.~\ref{fig:com:shape_scenario}(c). Thus, shape skeleton GT is generated by:
\begin{equation}
\small
\textbf{S}_{shape} = \mathtt{max}(p(\mathtt{min}(f(\textbf{S}^{\prime},\widehat{\textbf{M}})), \widehat{\textbf{M}}))
\quad .
\label{equ:per:shape}
\end{equation}
where $p(\cdot)$ and $f(\cdot)$ are same to Eq.~\ref{equ:per:image}. An intuitive example is presented in Fig.~\ref{fig:com:trade_off}. We can clearly observe the changes in simplicity (SS) and reconstruction error (RE) during the pruning process. With the hints from RE and SS, most volunteers tend to select the third one (marked by the rectangle) since it strikes the best balance, being structurally complete and relatively simple. As shown in Fig.~\ref{fig:com:image_scenario} (f) and Fig.~\ref{fig:com:shape_scenario} (d), skeleton GT generated by our strategies are perception-friendly, while at the same time properly balancing the skeleton simplicity and the shape reconstruction error.
\section{Annotation Tool and Ground Truth}
\label{s:tool}
In this section, we first introduce the design of an annotation tool, SkeView, based on our proposed strategy. Then, using SkeView, we generate GTs for the 17 existing datasets shown in Table~\ref{tab:intro:compare}.

\subsection{SkeView}
\label{s:tool:skeview}
\begin{figure}[t!]
  \centering
    \includegraphics[width=1\linewidth]{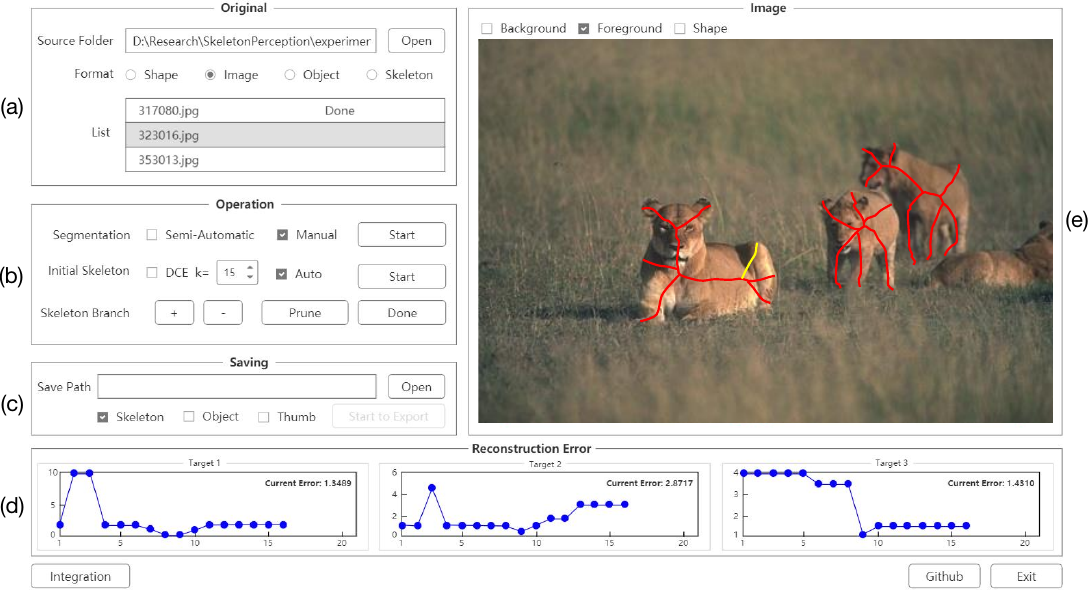}
  \caption{User interface of SkeView. (a) Data and format selection. (b) Operations including segmentation, initial skeleton generation and pruning. (c) Result format and exporting. (d) Reconstruction error and log. (e) Preview of image, object, shape and their skeletons.}
\label{fig:com:skeview}
\vspace{-0.5em}
\end{figure}

To facilitate the strategies in Eq.~\ref{equ:per:image} and~\ref{equ:per:shape}, we developed a tool, SkeView, for extracting skeleton GTs in shape and image datasets. The user interface contains five major panels (see Fig.~\ref{fig:com:skeview}):

\noindent{\bf (a) Source}. SkeView supports five source data types including shape, image, object (segmented foreground) and skeleton (only for pruning-related operations).

\noindent{\bf (b) Operations}. This includes image segmentation (only available for the ``Image" format), initial skeleton generation/selection, and dynamic skeleton branch pruning. For instance, there are two modes to address segmentation: manual and semi-automatic. If the manual mode is selected, users can dynamically plot a polygon to crop the region-of-interest. Otherwise, a Mask RCNN model~\cite{He2017MRC}, pre-trained with COCO dataset~\cite{Lin2014MCC}, is loaded to extract the initial segmentation masks. Then, the selected mask is transformed into a polygon by uniformly inserting interactive plots along the mask boundary. This way, each interactive plot can be manually moved to optimize the shape of the mask. For images with multiple objects (\textit{i.e.} SYMMAX300~\cite{Tsogkas2012LSD} and SymPASCAL~\cite{Ke2017SSR}), users can flexibly add and remove targets via buttons.

For initial skeleton extraction, the automatic mode~\cite{Shen2013SPA} is selected by default. According to the proposed strategy in Section~\ref{s:com:strategy}, we intentionally added slightly more branches in the initial skeleton to provide more flexibility in the following pruning step. SkeView also allows users to generate initial skeleton semi-automatically using the discrete curve evolution (DCE) method~\cite{Bai2007SPB} by varying the stop parameter $k$. Either way, users can coarsely add (or remove) skeleton branches by simply clicking on the ``+" (or ``-") buttons until the generated skeleton is satisfactory. SkeView preserves all branches in each step of the skeleton evolution from complex to simple in~\cite{Shen2013SPA,Bai2007SPB}. This operation is functionally similar to the skeleton selection process in Fig.~\ref{fig:com:image_scenario} (d). Finally, as presented in Fig.~\ref{fig:com:skeview} (e), users can finely prune redundant branches by selecting a target branch (marked in yellow) and clicking the ``Prune" button (or the ``Delete" key).

\noindent{\bf (c) Exports}. Each export format is a structure with multiple elements: ``Skeleton" (skeleton binary matrix, list of endpoints and junction points), ``Object" (segmented foreground, shape and boundary matrices) and ``Thumb" (pure skeleton and preview images, as shown in (e)). SkeView also preserves the pruning parameters and the correlated skeletons for future domain mapping and learning.

\noindent{\bf (d) Reconstruction error}. In this panel, current and historic reconstruction errors (Eq.~\ref{equ:per:error}) of each target are displayed during skeleton initialization and pruning. To facilitate comparison between the current and the previously pruned skeleton, the current reconstruction error is presented in bold font at the top right corner, and also plotted dynamically (as blue points) on the graph. Moreover, users can easily click a point to load the previous pruning result for visualization and reconsideration.

\noindent{\bf (e) Preview and branch selection}. Users can preview images, segmented objects and initial skeletons in this panel. Similar to the APP in Fig.~\ref{fig:com:app}, the background transparency, skeleton colour and boundary visibility can be adjusted here. During the fine-grained pruning process, users can select multiple branches by clicking on the target while pressing the ``Shift" key.

\begin{figure}[t!]
  \centering
    \includegraphics[width=1\linewidth]{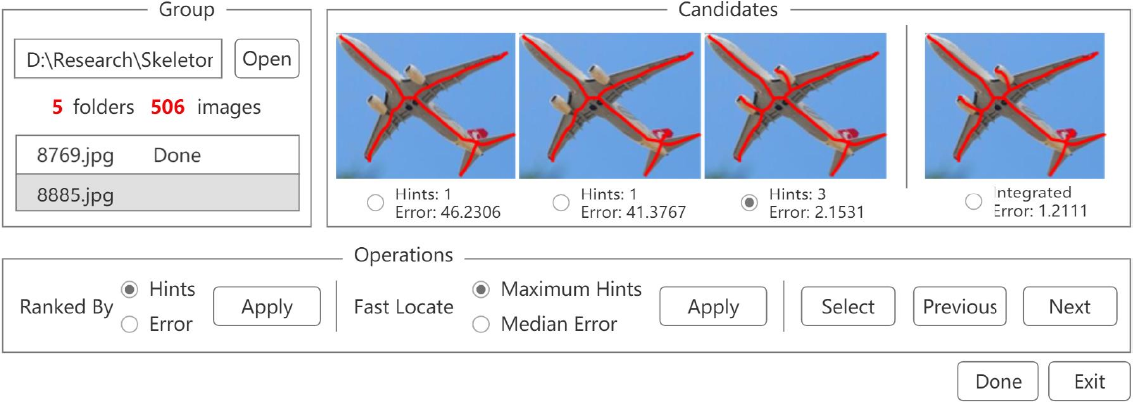}
  \caption{User interface of the Integration Function.}
\label{fig:com:integration}
\vspace{-0.5em}
\end{figure}
Tsogkas {\it et al.}~\cite{Tsogkas2016MRF} have introduced a tool with a user interface for annotating skeletons by manually drawing poly-lines. Besides being less efficient due to its purely manual operation, it also cannot ensure the symmetry of poly-lines according the 2D object contour. SkeView is advantageous in both these aspects. As SkeView is developed for individual users, we also provide a tool for skeleton integration and selection from a group of users (Fig.~\ref{fig:com:skeview} (bottom left)). As presented in Fig.~\ref{fig:com:integration}, skeletons from multiple users are presented together for final determination of the acceptable annotation. The tool can automatically count the duplicated skeletons (``Hints") and calculate reconstruction error (``Error"). By default, the final skeleton is automatically selected according to the maximum ``Hints" and median ``Error". For groups with fewer than three volunteers, SkeView integrates branches from the candidates to extract a new skeleton candidate. To evaluate the efficiency, we compare SkeView with the method in~\cite{Shen2016OSE} on SK506 dataset. Our statistics show that the time cost per image is reduced from 86.4 to 27.2 seconds. This suggests that SkeView is suitable for medium-scale datasets which are mostly those listed in Table~\ref{tab:intro:compare}.

For large-scale datasets ({\it e.g.} more than 10,000 shapes), an efficient way is to generate initial skeletons using the SkeView semi-automatic method with big pruning power ({\it e.g.} $k=50$). The pruning process is conducted via online labelling tools (such as LabelMe~\cite{Russell2008LAD}) by drawing bounding boxes on the endpoints that are intended to preserve. The pruned skeleton is generated by mapping skeleton paths between the preserved endpoints to a zero matrix. Compared with the branch-based pruning in Fig.~\ref{fig:com:skeview} (c), the box-based pruning only offers a slight consideration of the context arising from dense endpoints. However, it is more efficient for the purpose of group collaboration with its range of rich online annotation tools~\cite{Dasiopoulou2011ASO}. SkeView is developed with Matlab R2015b with GUIDE for user interface. The toolbox SkeView is compiled into executable applications in both Windows and Linux. The source codes and datasets are publicly available in this repository\footnote{\url{https://github.com/cong-yang/skeview }}.

\subsection{Ground Truth}
\label{s:tool:gt}
To ensure the quality of annotation, each GT was generated by four participants (two males and two females) from NENU. To meet different requirements in image and shape datasets, image GTs included segmented foregrounds, binary shapes, skeletons, lists of endpoints and junction points. Shape GTs included skeletons, and the list of endpoints and junction points. All skeleton branches in our GTs are one pixel wide, and are connected to shape boundaries: this meets the quality requirements of most skeleton extraction and matching algorithms. Users can intentionally dilate and dilute a GT skeleton point depending on algorithms~\cite{Wang2019DFS,Atienza2019PUF}. In practice, there are two strategies to ensure the application requirements on GT properties:
\begin{itemize}
    \item Annotation documents: written by domain experts, detail the annotation and quality requirements, including annotation examples and corner cases.
    \item Annotation training: annotators (volunteers in our case) study the annotation documents, followed by a trial-checking process using some samples. 
\end{itemize}
Built on that, annotators not only follow their perception of skeleton simplicity and reconstruction error, but also consider the requirements of different domains. Besides, such strategies can ensure the quality and consistency of GTs. In our case, the extracted shape skeletons on the existing ten datasets are general enough for CNN-based skeleton detector training and skeleton matching. This is because these datasets were typically collected for the shape retrieval scenario. In terms of the four image datasets, the datasets are used for general object detection and analysis. Thus, our GTs not only respect their original setting and domain requirements, but also have better quality, clarity, and consistency.

\subsubsection{Image Datasets}
\label{s:gt:gt:image}
\begin{figure}[t!]
  \centering
    \includegraphics[width=0.7\linewidth]{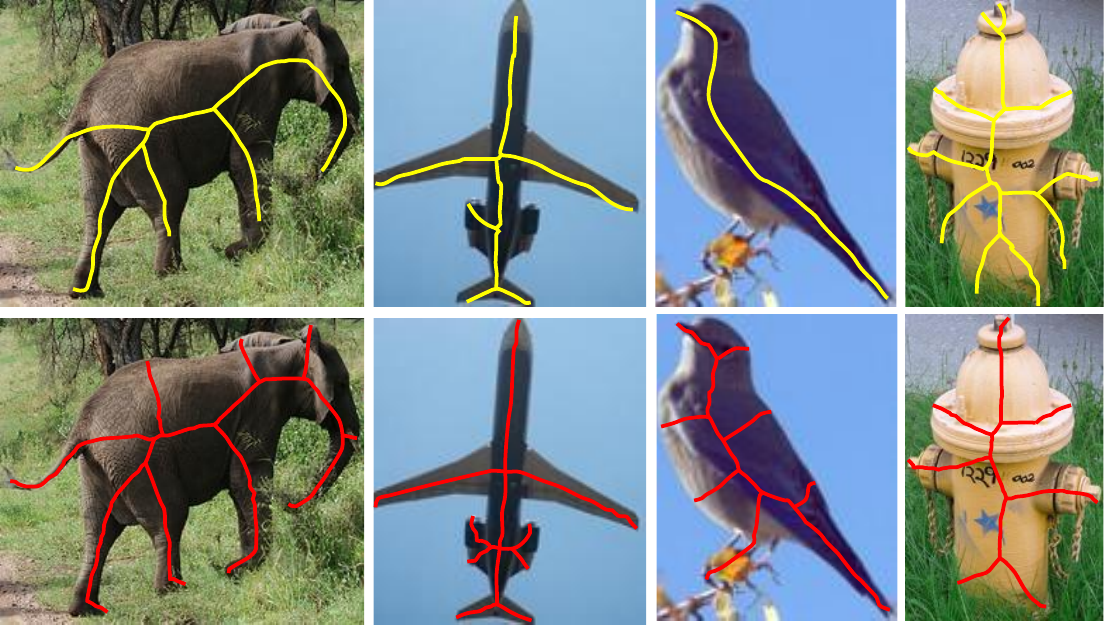}
  \caption{Comparison of the original (yellow) GTs with the ones generated with our SkeView (red) in SK506~\cite{Shen2016OSE} and SK1491~\cite{Shen2017DLM} datasets.}
\label{fig:gt:sk}
\vspace{-0.5em}
\end{figure}
\begin{figure}[t!]
  \centering
    \includegraphics[width=0.7\linewidth]{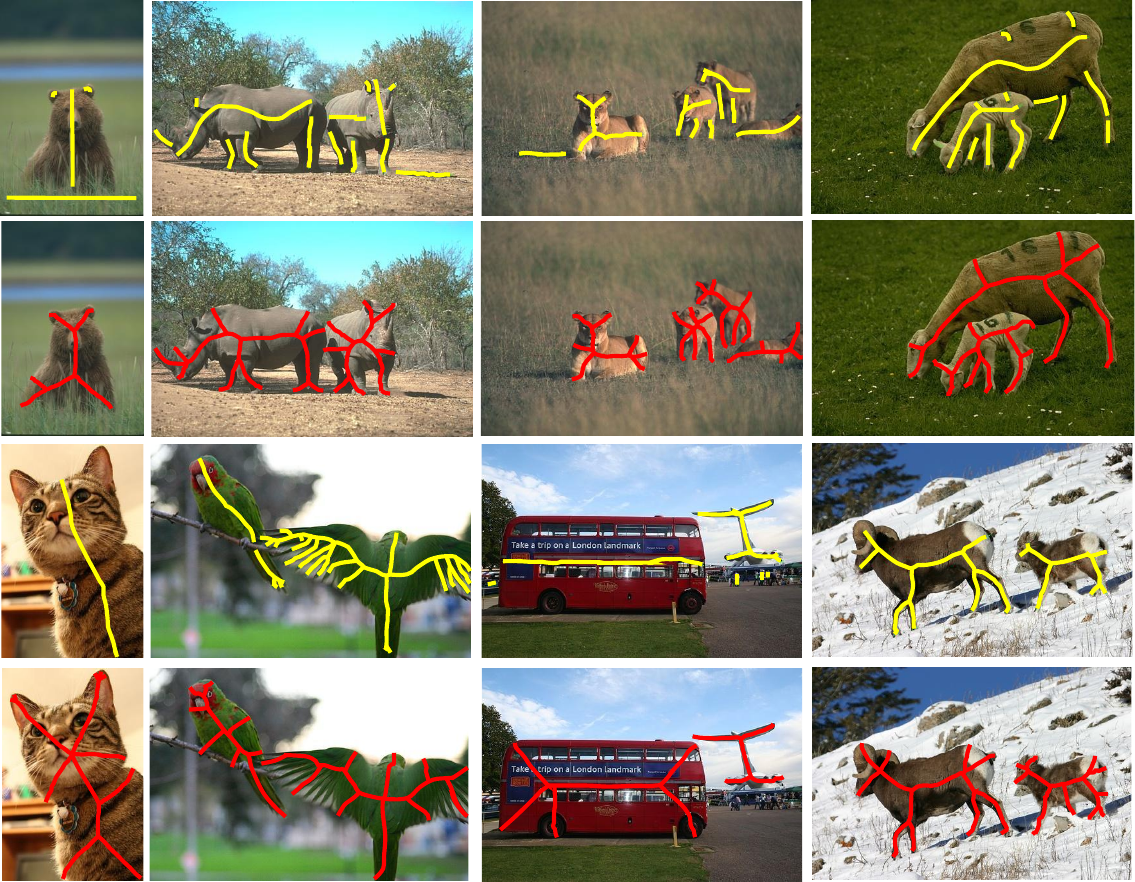}
  \caption{Comparison of the original (yellow) GTs with the ones generated with our SkeView (red) in SYMMAX300~\cite{Tsogkas2012LSD} (top two rows) and SymPASCAL~\cite{Ke2017SSR} (bottom two rows).}
\label{fig:gt:symmax}
\vspace{-0.5em}
\end{figure}
Fig.~\ref{fig:gt:sk} and~\ref{fig:gt:symmax} present comparisons between the original GTs and our GTs generated by SkeView among four image datasets:

\noindent{\bf SK506}~\cite{Shen2016OSE} (also known as SK-SMALL) was selected from the MS COCO~\cite{Lin2014MCC} dataset, with 506 natural images (300 for training and 206 for testing) and 16 object classes including humans, animals and artifacts. For each image, there is only one target for the skeleton GT generation. Due to inaccurate segmentation and unstable individual perception, the quality of the original GT is not promising. For instance, as shown in Fig.~\ref{fig:gt:sk} (top), we observe the following issues: shortened branches of the elephant, the asymmetric branches in the airplane, an overly-simplified skeleton for the bird, and noisy branches in the hydrant. In contrast, GTs generated by SkeView are better in terms of various qualitative properties: consistency, perception friendliness, and the representation of object geometrical features.

\noindent{\bf SK1491}~\cite{Shen2017DLM} (also known as SK-LARGE) is an extension of the SK506 by selecting more images from the MS COCO dataset. It includes 1,491 images (746 for training and 745 for testing). Similar to SK506, there is one target for each image and the GT skeletons are annotated in the same way. 

\noindent{\bf SYMMAX300}~\cite{Tsogkas2012LSD} is adapted from the Berkeley Segmentation Dataset (BSDS300)~\cite{Martin2001ADO} with 300 images (200 for training and 100 for testing). There are multiple targets in most images. This dataset is used for local reflection symmetry detection, which is a low-level image feature, without paying attention to the concept of `object'. While most branches are disconnected and the original GTs do not encode information about the connectivity of skeleton branches. Hence, it is ill-suited to evaluate object skeleton extraction methods as a large number of symmetries occur in non-object parts (see the bear, rhinoceros and lion images in Fig.~\ref{fig:gt:symmax} (top)). For this, we regenerated GTs only on target objects, as it was more meaningful to use object symmetry (foreground) instead of whole-image symmetry. As suggested in~\cite{Tsogkas2016MRF}, we ignore images without specific target objects.

\noindent{\bf SymPASCAL}~\cite{Ke2017SSR} was selected from the PASCAL-VOC dataset~\cite{Everingham2010TPV}, with 1,435 images (648 for training and 787 for testing). Most images contain multiple targets, partial visibility and complex backgrounds. However, there are still noisy symmetries from the background, incomplete skeleton graph and shortened skeleton branches. In contrast, GTs from SkeView focus only on the foregrounds, maintaining the same quality as with the other three image datasets. In Fig.~\ref{fig:gt:symmax}, we clearly observe that our GTs in SYMMAX300 and SymPASCAL have the same quality as SK506, and skeleton branches for each object are well-connected. Such features can ensure a reliable evaluation on both skeleton extraction and matching algorithms~\cite{Bai2008PSS}.

It should be noted that our annotation mainly captures the 2D contours, and partly loses the 3D symmetry awareness for some objects in images. However, our labelling is superior to the original GTs of the four image datasets, especially considering the consistency standards, branch connectivity and distinguished graphs. As a result, our GTs are more applicable for training and testing CNN-based skeleton detectors, as well as benchmarking skeleton-related pruning, matching and classification algorithms. In the future, we plan to update SkeView for 3D object and symmetry annotation~\cite{Tagliasacchi2016SAS} based on our strategy.

\subsubsection{Image and Shape Datasets}
\label{s:gt:gt:both}
\begin{figure}[t!]
  \centering
    \includegraphics[width=0.7\linewidth]{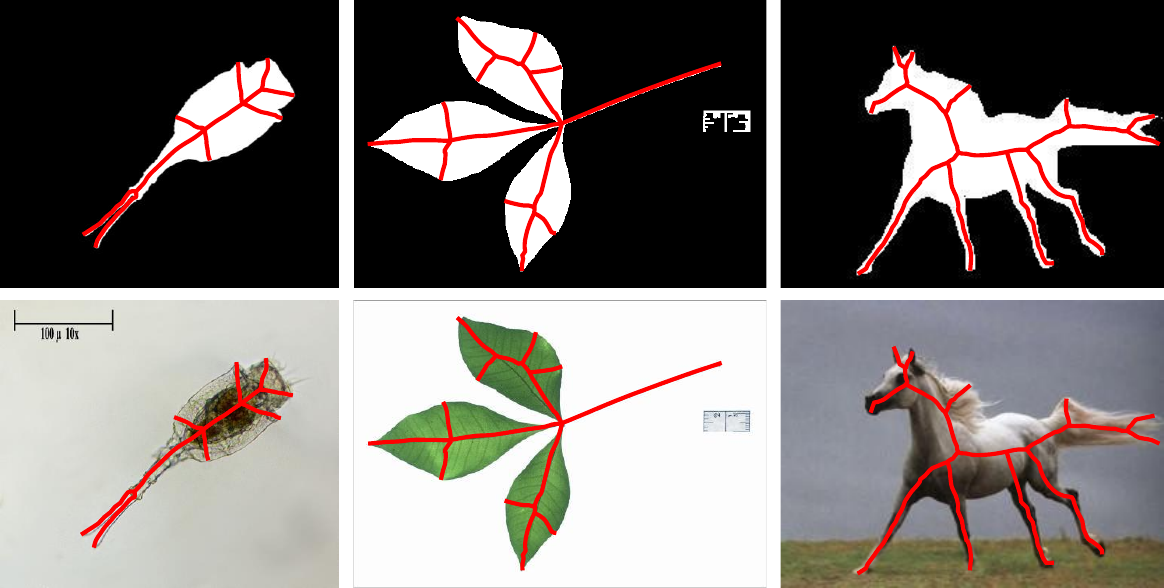}
  \caption{From left to right, skeleton GT of EM200~\cite{Yang2014SCO}, SmithsonianLeaves~\cite{Haibin2007SCU} and WH-SYMMAX~\cite{Shen2016MIS} generated by SkeView.}
\label{fig:gt:both}
\vspace{-0.5em}
\end{figure}
There are three datasets with both images and corresponding foreground shapes (Fig.~\ref{fig:gt:both}). For this, we extracted initial skeletons using shapes and applied pruning using images in SkeView.

\noindent{\bf EM200}~\cite{Yang2014SCO} contains 200 microscopic images (10 classes) of environmental microorganisms (EM). There are two types of segmented foregrounds provided by the original dataset: those generated manually or semi-automatically with the methods introduced in~\cite{Li2013AMA}. This dataset is challenging on colourless, transparent and spindly regions (flagellum). To ensure the quality of GTs, we employed the manual approach for initial skeleton generation. Then an efficient pruning in SkeView can best protect skeleton branches in those spindly regions for fine-grained EM matching and classification.

\noindent{\bf SmithsonianLeaves}~\cite{Haibin2007SCU} contains 343 leaves (187 for training and 156 for testing) from 93 different species of plants. Each leaf was photographed on a plain background. K-means clustering was employed to estimate the foreground based on colour, followed by morphological operations to fill in small holes. Thus, this dataset is relatively less challenging with respect to occlusion and complex backgrounds, but has richer geometrical characteristics. Our GTs can be used by botanists to compute leaf similarity in the digital archives of the specimen types.

\noindent{\bf WH-SYMMAX}~\cite{Shen2016MIS} contains 328 cropped images (228 for training and 100 for testing) from the Weizmann Horse dataset~\cite{Borenstein2002CTS}. Each image contains one manually segmented target. The original skeleton annotations are not only inconsistent concerning completeness across different horse shapes but also contain shortened branches. On the other hand, our GTs yield better quality with respect to consistency and completeness.

\subsubsection{Shape Datasets}
\label{s:gt:gt:shape}
\begin{figure}[t!]
  \centering
    \includegraphics[width=1\linewidth]{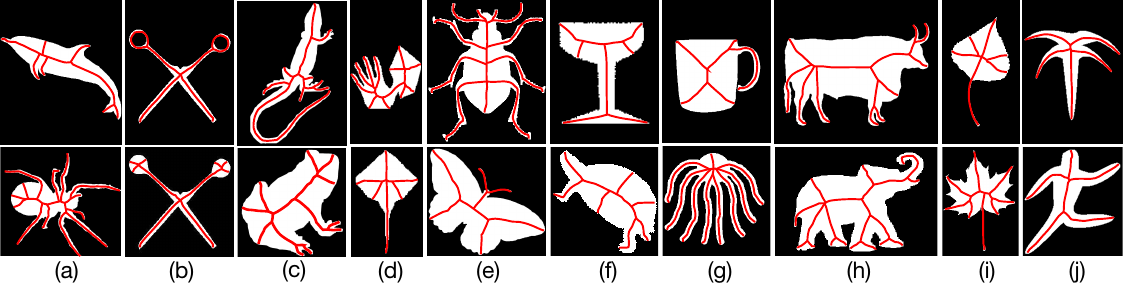}
  \caption{Skeleton GTs of (a) Animal2000~\cite{Bai2009ICA}, (b) ArticulatedShapes~\cite{Haibin2007SCU}, (c) SkelNetOn~\cite{Ilke2019SDA}, (d) Kimia99~\cite{Sebastian2004ROS}, (e) MPEG7~\cite{Latecki2000SDF}, (f) Kimia216~\cite{Sebastian2004ROS}, (g) MPEG400~\cite{Cong2014SBO}, (h) Tetrapod120~\cite{Cong2016OMW}, (i) SwedishLeaves~\cite{Soderkvist2001CVC}, (j) Tari56~\cite{Asian2005AAR} datasets generated by SkeView.}
\label{fig:gt:shapes}
\vspace{-0.5em}
\end{figure}
Fig.~\ref{fig:gt:shapes} presents samples of ten shape datasets and their GTs generated by SkeView:

\noindent{\bf Animal2000}~\cite{Bai2009ICA} contains 2,000 shapes (20 categories, 100 shapes each) ranging from poultry and domestic pets to insects and wild animals. Each class is characterised by large intra-class shape variations. Due to occlusion, some parts of certain objects ({\it e.g.} legs) are missing. There are also holes and boundary noises in some shapes due to incorrectly segmented foregrounds and backgrounds. This dataset is actively used in shape matching, classification and skeleton-based shape retrieval. As the shape category can be easily identified by human perception, critical parts of an object ({\it e.g.} legs, head, tentacles) are all preserved by the skeleton branches of our GTs.

\noindent{\bf ArticulatedShapes}~\cite{Haibin2007SCU} contains 40 images from eight different objects. This challenging dataset consists of various tools including scissors with holes. To preserve the original topology, our GTs at such regions are closed branches (Fig.~\ref{fig:gt:shapes} (b) (top)). Most existing matching algorithms~\cite{Bai2008PSS} cannot properly deal with skeleton graph structures with cycles, however we could provide skeleton GTs after filling the holes (Fig.~\ref{fig:gt:shapes} (b) (bottom)). 

\noindent{\bf SkelNetOn}~\cite{Ilke2019SDA} contains 1,725 shapes (1,218 for training, 241 for validation and 266 for testing) represented as pixels. All shapes are of high quality with the holes and isolated pixels having removed by morphological operations (dilation and erosion) and manual adjustments. However, skeleton branches in this dataset are shortened and suffer from imbalance in simplicity, {\it i.e.} the original GTs in some shapes are extremely simple while others are overly complex. As such, it is difficult to conduct a fair comparison on skeleton-related algorithms such as extraction and matching. Moreover, this dataset is available only to registered participants in the SkelNetOn Challenge~\cite{Ilke2019SDA}. Our GTs are only for the purpose of skeleton quality analysis as shown in Table~\ref{tab:exp:gt:std}. 

\noindent{\bf Kimia99}~\cite{Sebastian2004ROS} contains 99 shapes (9 categories, 11 shapes each) assembled from a variety of sources such as tools and hands, etc. Challenges in each category come from occlusion, and articulation of missing parts. To avoid topology violation of shapes, branches of extrinsic regions ({\it e.g.} Fig.~\ref{fig:gt:shapes} (d) (top)) are preserved in GTs. 

\noindent{\bf MPEG7}~\cite{Latecki2000SDF} contains 1,400 (70 categories, 20 shapes each) shapes defined by their outer closed contours. It poses challenges with respect to deformation ({\it e.g.} change of view points and non-rigid object motion) and noises ({\it e.g.} quantisation and segmentation noise). This dataset is actively used for benchmarking shape representation, matching and retrieval algorithms~\cite{Cong2016OMW,Yang2020TAS}. Similar to Kimia99, our GTs respect the topology of original shapes and properly preserve the challenges posed in each category.

\noindent{\bf Kimia216}~\cite{Sebastian2004ROS} contains 216 shapes (18 categories, 12 shapes each) selected from the MPEG7 dataset. It is actively used in skeleton extraction, pruning, matching and shape retrieval scenarios. Our GTs in this dataset form a subset of MPEG7.

\noindent{\bf MPEG400}~\cite{Cong2014SBO} contains 400 shapes selected from the MPEG7 dataset (20 categories, 20 shapes each). Instead of directly using the original shapes, boundary noises of these shapes were manually removed for ablation study. Thus, our GTs are slightly different from the corresponding ones in the MPEG7 dataset.

\noindent{\bf Tetrapod120}~\cite{Cong2016OMW} contains 120 tetrapod animal shapes from six classes. As shapes of some species are visually similar, this dataset is normally employed to evaluate shape matching and fine-grained classification algorithms. An advantage of SkeView is that branches of major regions are preserved. However, our GTs are not recommended for evaluating fine-grained classification algorithms as some animal species can only be distinguished via branches in small regions ({\it e.g.} floppy vs. pointy ears). 

\noindent{\bf SwedishLeaves}~\cite{Soderkvist2001CVC} contains 1,125 leaf shapes from 15 different Swedish tree species, with 75 leaves per species (25 for training, 50 for testing). This dataset is challenging as some species are quite similar. Past works~\cite{Soderkvist2001CVC,Haibin2007SCU} have shown that it is not possible to distinguish them based on shape features alone. We do not intend to perform the same task using our GT skeletons. Instead, our GTs can be used for a wider scope of tasks -- evaluating general skeleton extraction, pruning and matching algorithms.

\noindent{\bf Tari56}~\cite{Asian2005AAR} contains 56 shapes (14 categories, 4 shapes each) for evaluating matching performance under visual transformations. Shapes of the same category show variations in orientation, scale, articulation and small boundary details. Motivated by this, our GT skeletons are useful for evaluating various skeleton-based shape matching algorithms. This is because our GTs contain branches with respect to the major and contextual shape regions. Moreover, our skeleton GTs are inherently robust to orientation and scale.

\subsubsection{Properties}
\label{s:tool:properties}
We discuss two measured properties of skeleton GTs: the mean Reconstruction Error (RE) and Skeleton Simplicity (SS). RE is already calculated by Eq.~\ref{equ:per:error}. Here, SS is calculated by:
\begin{equation}
\small
s(\textbf{S}) = \mathtt{exp}(-\mathtt{log}(\Gamma(S)+1))
\quad .
\label{exp:simplicity}
\end{equation}
where $\Gamma(S)$ denotes the normalized curve length of skeleton $S$. Since the GT skeletons are one pixel wide, $\Gamma(S)$ can be calculated simply from the number of skeleton points, normalized by the average path length of the skeleton. A constant value of 1 is added to ensure that the value from log function is positive. Eq.~\ref{exp:simplicity} is motivated by the intuitive understanding that shorter skeletons have simpler structures. Here, SS is used for quantitative analysis since the differences between GTs are primarily at the fine-grained level. We note that another quantitative way to measure the simplicity of $\textbf{S}$ is to use the number of junction and endpoints. However, experiments in~\cite{Yang2020TAS} show that endpoint statistics (both mean and standard deviation values) from different methods are similar to each other. In contrast, $\Gamma(S)$ is more distinguishable as it is sensitive to slight changes in the skeleton structure. In other words, SS has higher discriminative power than the number of endpoints, particularly at the fine-grained level.

\begin{table}[t!]
\centering
\setlength{\tabcolsep}{3.7pt}
\caption{Mean reconstruction error (RE) and skeleton simplicity (SS) of our GTs. Skeletons from the automatic approaches DCE~\cite{Bai2007SPB} (fixed $k=10$), AutoSke~\cite{Shen2013SPA} and Grafting~\cite{Yang2020TAS} are also detailed for reference. In each dataset, the lowest RE (towards complete skeleton structure) and the highest SS (towards simple skeleton structure) values are in boldface.}
\label{tab:exp:gt:std}
\renewcommand{\arraystretch}{1.1}
\setlength{\arrayrulewidth}{1pt}
\begin{tabular}{lcccccccccccc}
\hline
 & \multicolumn{2}{c}{SK1491} & \multicolumn{2}{c}{EM200} & \multicolumn{2}{c}{Kimia216} & \multicolumn{2}{c}{MPEG7} & \multicolumn{2}{c}{Animal2000} & \multicolumn{2}{c}{SkelNetOn} \\
& RE & SS & RE & SS & RE & SS & RE & SS & RE & SS & RE & SS \\ \hline
DCE & 0.90 & \textbf{0.09} & \textbf{0.90} & 0.07 & \textbf{0.81} & 0.09 & \textbf{0.92} & 0.08 & 0.86 & \textbf{0.09} & 0.87  & 0.07 \\
AutoSke & 0.89 & 0.08 & 0.91 & \textbf{0.15} & 0.82 & \textbf{0.11} & \textbf{0.92} & \textbf{0.11} & 0.86 & \textbf{0.09} & 0.85 & \textbf{0.09} \\
Grafting & 0.89 & 0.07 & \textbf{0.90} & 0.13 & 0.82 & \textbf{0.11} & \textbf{0.92} & \textbf{0.11} & 0.86 & 0.08 & 0.85 & \textbf{0.09} \\ 
GTs & \textbf{0.88} & 0.05 & \textbf{0.90} & 0.07 & \textbf{0.81} & 0.08 & \textbf{0.92} & 0.08 & \textbf{0.85} & 0.07 & \textbf{0.82} & 0.07 \\ \hline
\end{tabular}
\vspace{-0.5em}
\end{table}
Table~\ref{tab:exp:gt:std} presents the RE and SS of our GTs. For comparison, skeletons generated by three automatic approaches DCE~\cite{Bai2007SPB} (with the fixed stop parameter $k=10$ recommended in~\cite{Cong2016IOS}), AutoSke~\cite{Shen2013SPA} and Grafting~\cite{Yang2020TAS} are also presented. We first report their statistical distribution of RE and SS values. Taking the Kimia216 dataset as an example, the statistics of 216 shapes are twofold: statistics within the same method and between different methods:
\begin{itemize}
    \item With our proposed SkeView, the values are close to each other. Notably, the statistical distribution of RE is between 0.80-0.82, and SS is between 0.08-0.09. In other words, our GTs strike a stable balance, being structurally complete and relatively simple.
    \item With different approaches (DCE, AutoSke, Grafting), the distributions are more varied. RE and SS of DCE: 0.73-0.97, 0.05-0.09; RE and SS of AutoSke: 0.67-0.94, 0.06-0.14; RE and SS of Grafting: 0.73-0.91, 0.06-0.14. Though the mean values of RE and SS are close to SkeView, skeleton structures are unstable. In other words, some skeletons are either too simple, or too complex. This phenomenon is inherently similar to our observations in Section~\ref{s:com:dia}, such as O1 (Perception is robust to the time and volunteer groups) and O2 (Perception is robust to segmented objects and images).
\end{itemize}

\begin{figure}[t!]
  \centering
    \includegraphics[width=0.6\linewidth]{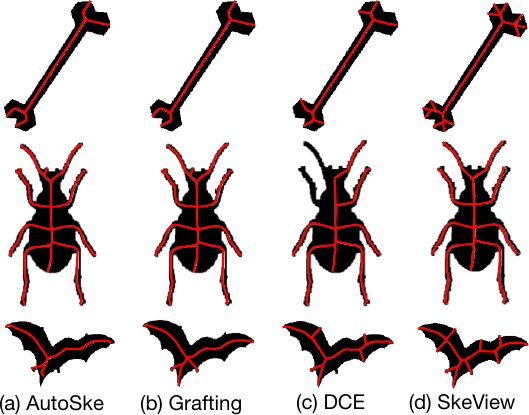}
  \caption{Sample GTs generated by (a) AutoSke~\cite{Shen2013SPA}, (b) Grafting~\cite{Yang2020TAS}, (c) DCE~\cite{Bai2007SPB} and (d) SkeView in MPEG7 dataset.}
\label{fig:gt:compare_gt}
\vspace{-0.5em}
\end{figure}
We also observe that our GTs have the lowest RE, while the structures are more complex (smaller SS means more complex structure). For instance, RE in the MPEG7 datasets are the same (0.92), while SS of AutoSke~\cite{Shen2013SPA} and Grafting~\cite{Yang2020TAS} are the smallest (0.11). However, as shown in Fig.~\ref{fig:gt:compare_gt} (a) and (b), their skeleton completeness are not perceptually promising. Though skeletons generated by the DCE method~\cite{Bai2007SPB} are the simplest in both SK1491 and Animal2000 datasets, their RE are relatively high (the first row in Table~\ref{tab:exp:gt:std}) while their skeleton structures are not visually promising (Fig.~\ref{fig:gt:compare_gt} (c)). Overall, our GT skeletons strike the best balance, being perceptually friendly, structurally complete (in most cases), and relatively simple (the median SS is only 0.03 lower than AutoSke).
\section{Benchmarks}
\label{s:bm}
In this section, we present a benchmark evaluation of skeleton detectors (mostly CNN-based methods) and skeleton-based matching methods using our GTs. For fairness, all settings follow their original papers unless stated otherwise.

\subsection{Skeleton Detectors in Shapes}
\label{s:gt:baselines:shape}
\begin{figure}[t!]
  \centering
    \includegraphics[width=0.85\linewidth]{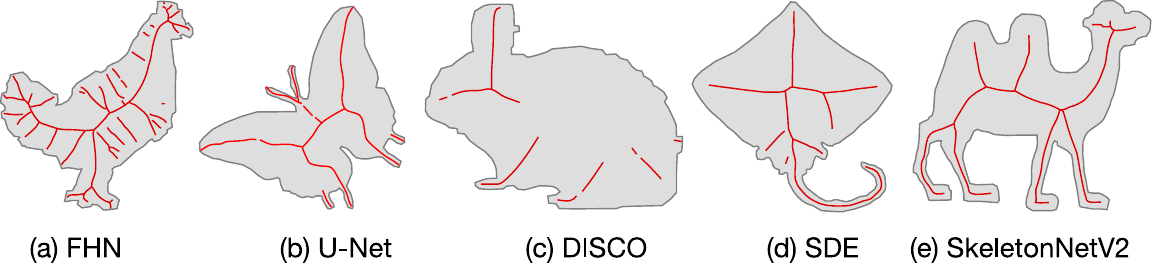}
  \caption{Shape skeleton extraction using CNN-based methods: FHN~\cite{Jiang2019FHN}, U-Net~\cite{Panichev2019}, DISCO~\cite{Song2021DUB}, SDE~\cite{Tang2021DAE}, and SkeletonNetV2~\cite{Nathan2021SAD}.}
\label{fig:exp:shapeskeletons}
\end{figure}

\begin{table}[t!]
\centering
\setlength{\tabcolsep}{5.5pt}
\caption{Average error pixel (AEP) of shape skeletons from different methods and datasets. Ani2000: Animal2000. SL1: SmithsonianLeaves. SL2: SwedishLeaves. AS: ArticulatedShapes. EM: EM200. The smallest AEP in each dataset are shown in boldface.}
\label{tab:exp:baseline:aep}
\renewcommand{\arraystretch}{1.1}
\setlength{\arrayrulewidth}{1pt}
\begin{tabular}{lcccccccc}
\hline
 & Kimia216 & Ani2000 & SL1 & SL2 & Tari56 & MPEG7 & AS & EM  \\ \hline
DCE~\cite{Bai2007SPB} & 1.04 & \textbf{0.97} & 8.05 & 5.84 & 0.91 & 3.90 & 0.61 & 6.18 \\
AutoSke~\cite{Shen2013SPA} & \textbf{0.80} & 1.01 & \textbf{3.67} & \textbf{3.19} & \textbf{0.51} & \textbf{3.03} & \textbf{0.39} & \textbf{4.10} \\
Physics~\cite{Krinidis2009ASF} & 1.29 & 1.18 & 10.15 & 7.09 & 1.09 & 4.73 & 0.67 & 7.47 \\
BPR~\cite{Shen2011SGA} & 0.88 & 1.14 & 4.13 & 3.66 & 0.56 & 3.33 & 0.44 & 4.55 \\
U-Net~\cite{Panichev2019} & 1.41 & 1.32 & 11.15 & 7.87 & 1.32 & 5.65 & 0.81 & 8.32 \\ \hline
\end{tabular}
\vspace{-0.5em}
\end{table}
To quantitatively evaluate the performance of different skeleton detectors, we employed the average error pixel (AEP) proposed in~\cite{Krinidis2009ASF} as the error measure. Specifically, it measures the error $e(\widehat{\textbf{S}}, \textbf{S})$ between a detected skeleton $\widehat{\textbf{S}}$ against a GT $\textbf{S}$ using the mean square error of their skeleton points:
\begin{equation}
\small
e(\widehat{\textbf{S}}, \textbf{S}) = \frac{1}{N}\sum_{i=1}^{N}(\sqrt[]{(\widehat{\textbf{S}}_{x}(i)-\textbf{S}_{x}(i))^{2}+(\widehat{\textbf{S}}_{y}(i)-\textbf{S}_{y}(i))^{2}}
\quad .
\label{exp:aep}
\end{equation}
where $(\widehat{\textbf{S}}_{x}(i), \widehat{\textbf{S}}_{y}(i))$ are the coordinates of a skeleton point in $\widehat{\textbf{S}}$, $N$ is their total number of points, and $(\textbf{S}_{x}(i)), \textbf{S}_{y}(i))$ is the closest point in $\textbf{S}$ to the point $(\widehat{\textbf{S}}_{x}(i), \widehat{\textbf{S}}_{y}(i))$. Table~\ref{tab:exp:baseline:aep} details the evaluation results of five representative methods on eight shape datasets. The Physics method~\cite{Krinidis2009ASF} generates skeleton points iteratively starting from a boundary point set based on a physics-based deformable model. Though it can be used to obtain stable skeletons with a fixed parameter setting, the results are not symmetric to the boundary and are sensitive to noises. The BPR~\cite{Shen2011SGA} method of pruning skeletons is based on the context (modelled by the bending potential ratio) of the boundary segment that corresponds to the branch. The U-Net~\cite{Panichev2019} is a typical CNN-based method, which employs a modified U-Net architecture for direct skeleton regression. We can clearly see that most of the skeletons generated by AutoSke~\cite{Shen2013SPA} have the lowest AEP and thus are closest to GTs. Though the DCE~\cite{Bai2007SPB} method achieves the best result on Animal2000, it is still close to the result generated by AutoSke (only around 0.04 lower). Among all the methods, the CNN-based U-Net has the lowest performance, with around 0.31 and 2.62 higher AEP than the AutoSke method in Animal2000 and MPEG7, respectively. This is because skeletons generated by CNN-based methods normally yields low-quality branches~\cite{Yang2022blumnet}. To verify it, we visualize skeletons from existing five CNN-based methods trained using our GTs (see Fig.~\ref{fig:exp:shapeskeletons}). We can clearly observe the noisy, disjointed, and incomplete skeleton branches.

\subsection{Skeleton Detectors in Images}
\label{s:gt:baselines:images}
\begin{table}[t!]
\centering
\setlength{\tabcolsep}{7.5pt}
\caption{F1 scores of skeleton detectors in images. SYMM: SYMMAX300. SymP: SymPASCAL. SL: SmithsonianLeaves. WHS: WH-SYMMAX.}
\label{tab:exp:baseline:fmeasure}
\renewcommand{\arraystretch}{1.1}
\setlength{\arrayrulewidth}{1pt}
\begin{tabular}{lccccccc}
\hline
 & SK506 & SK1491 & SYMM & SymP & EM200 & SL & WHS \\ \hline
HED~\cite{Xie2015HED} & 0.552 & 0.494 & 0.431 & 0.370 & 0.298 & 0.580 & 0.741 \\
SRN~\cite{Ke2017SSR} & 0.652 & 0.677 & 0.447 & 0.443 & 0.303 & 0.593 & 0.780 \\
Hi-Fi~\cite{Zhao2018HHF} & 0.693 & 0.727 & 0.460 & 0.458 & 0.311 & 0.620 & 0.822 \\
DeepFlux~\cite{Wang2019DFS} & 0.715 & 0.752 & 0.494 & \textbf{0.520} & 0.315 & 0.625 & 0.849 \\
Ada-LSN~\cite{Liu2020ALS} & \textbf{0.748} & \textbf{0.798} & \textbf{0.497} & 0.504 & \textbf{0.319} & \textbf{0.672} & \textbf{0.883} \\ \hline
\end{tabular}
\vspace{-0.6em}
\end{table}
\begin{figure}[t!]
  \centering
    \includegraphics[width=1\linewidth]{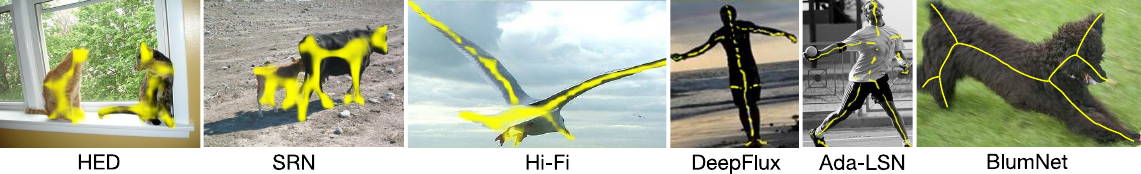}
  \caption{Image skeleton detection results from HED~\cite{Xie2015HED}, SRN~\cite{Ke2017SSR}, Hi-Fi~\cite{Zhao2018HHF}, DeepFlux~\cite{Wang2019DFS}, Ada-LSN~\cite{Liu2020ALS}, and BlumNet~\cite{Yang2022blumnet} methods, trained using our GTs.}
\label{fig:baseline:results:images}
\vspace{-0.6em}
\end{figure}

Skeletons in image are usually represented by binary maps after applying non-maximal suppression (NMS) and thresholding. The binary maps between the generated and the GT skeletons are matched pixel-wise to calculate the precision and recall values of the skeleton points. In practice, some small localization errors are allowed. Here, we used the F1 score ({\it i.e.} 2$\times$(precision$\times$recall)/(precision+recall)) to evaluate the performance of skeleton detectors on image datasets. Particularly, five CNN-based methods (HED~\cite{Xie2015HED}, SRN~\cite{Ke2017SSR}, Hi-Fi~\cite{Zhao2018HHF}, DeepFlux~\cite{Wang2019DFS} and Ada-LSN~\cite{Liu2020ALS}) are trained and tested using our GTs. To address a fair evaluation, we used their original settings on the backbone, image input size, and loss function. The initial network weights were initialized by Xavier. We optimized the loss functions using AdamW~\cite{Loshchilov2018DWD} with a mini-batch size of 1, an initial learning rate of 2e-4, and a weight decay of 1e-4. Regarding data augmentation, we applied random transformations to the image, including rotation, flipping, resizing and color jittering. For each method, we generated a precision-recall curve by varying the threshold value. The optimal threshold was selected as the one that produces the highest F1 score along the curve. 

Their best results are presented in Table~\ref{tab:exp:baseline:fmeasure}. We can see that Ada-LSN~\cite{Liu2020ALS} achieves the highest score on almost all datasets. Fig.~\ref{fig:baseline:results:images} presents some sample results. It can be observed that the predictions are generally convergent towards our GTs, though their performances are clearly different. Particularly, most methods are not feasible to generate high-quality skeleton graphs. For instance, skeletons from HED (Holistically-Nested Edge Detection), SRN (Side-output Residual Network), and Hi-Fi (Hierarchical Feature integration) contains lots of noise and limited smoothness. Deep-Flux and AdaLSN (Adaptive Linear Span Network) output clearer and slimmer results, while there remain some false positive points, disjointed segments, and incomplete branches. The main reason is that most CNN-based methods output noisy, disjointed, and incomplete skeleton branches in heat maps (also called skeleton maps). Some networks cannot guarantee the topological and geometrical features in the representation. In practice, skeleton heat maps from existing CNN-based methods usually require heavy and semi-automatic processes to extract slim skeletons (one pixel wide). Nevertheless, the geometrical and topological features of the processed skeletons are still not ensured. We can also observe that F1 scores on the EM200 are the lowest among all the evaluated datasets. A major reason for this occurrence is that the training data is very limited (only 10 images), resulting in under-fitted models.

It is still possible to extract one-pixel-wide skeleton graphs. As presented in Fig.~\ref{fig:baseline:results:images} (rightmost), we introduced a novel framework, BlumNet, for object skeleton extraction from shapes and images~\cite{Yang2022blumnet}. Unlike skeleton heat map regression with existing CNN-based methods, BlumNet decomposes a skeleton graph into structured components and simplifies the skeleton extraction problem into graph component detection and assembling tasks. Consequently, the quality of extracted skeletons is dramatically improved since BlumNet directly outputs slim, low-noise, and identified skeleton graphs. It should be noted that BlumNet was trained using our GTs from SkeView.

\begin{figure}[t!]
  \centering
    \includegraphics[width=0.9\linewidth]{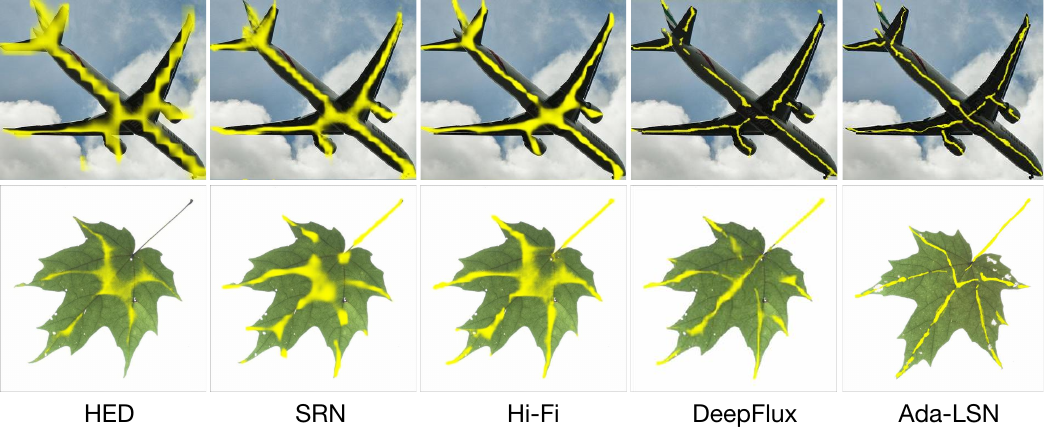}
  \caption{Visual comparison of HED~\cite{Xie2015HED}, SRN~\cite{Ke2017SSR}, Hi-Fi~\cite{Zhao2018HHF}, DeepFlux~\cite{Wang2019DFS}, and Ada-LSN~\cite{Liu2020ALS} skeletons, trained using our GTs.}
\label{fig:baseline:leaf:compare}
\vspace{-0.4em}
\end{figure}
\begin{table}[t!]
\centering
\setlength{\tabcolsep}{7pt}
\caption{Skeleton detection performance (F1 scores) comparison of 6 methods trained and tested using GTs from Original and SkeView on SK1491.}
\label{tab:exp:f1:compare}
\renewcommand{\arraystretch}{1.1}
\setlength{\arrayrulewidth}{1pt}
\begin{tabular}{lccccc}
\hline
 & HED~\cite{Xie2015HED} & SRN~\cite{Ke2017SSR} & Hi-Fi~\cite{Zhao2018HHF} & DeepFlux~\cite{Wang2019DFS} & Ada-LSN~\cite{Liu2020ALS} \\ \hline
GT (Original) & \textbf{0.497} & \textbf{0.678} & 0.724 & 0.732 & 0.786 \\
GT (SkeView) & 0.494 & 0.677 & \textbf{0.727} & \textbf{0.752} & \textbf{0.798} \\ \hline
\end{tabular}
\vspace{-0.6em}
\end{table}
Fig.~\ref{fig:baseline:leaf:compare} visually compares the detected skeletons on two sample images from SK1491 and SmithsonianLeaves. Specifically, skeletons from DeepFlux~\cite{Wang2019DFS} and Ada-LSN~\cite{Liu2020ALS} are slimmer and these two methods yield better performances in terms of continuity and completeness. Skeletons produced by HED are not well-integrated and are prone to noise. Though SRN and Hi-Fi yield clearer skeletons, they are not smooth and contain many false positive points. It is also interesting to find that most methods yield a better F1 score using the training and testing data from our GTs. For instance, the Fl score of DeepFlux~\cite{Wang2019DFS} method improved from 0.732 to 0.752 on the SK1491 dataset (Table~\ref{tab:exp:f1:compare}). This is because, comparing to the original GTs in the SK1491, our GTs are more consistent and possess better completeness in representing objects' geometrical features (see Fig.~\ref{fig:gt:sk}). As a result, these CNN-based models converge faster and generalize easier.

\subsection{Skeleton Matching}
\label{s:gt:baselines:matching}
\begin{figure}[t!]
  \centering
    \includegraphics[width=0.75\linewidth]{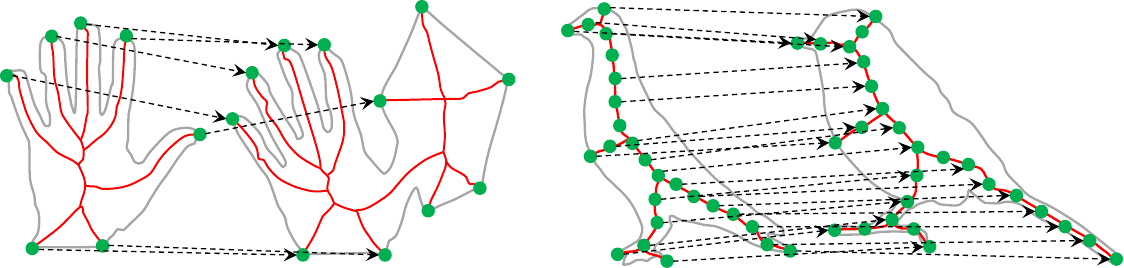}
  \caption{Skeleton-based shape matching algorithms in our evaluation. Left: Path Similarity (PS)~\cite{Bai2008PSS}. Right: Skeleton Context (SC)~\cite{Kamani2016SMU}.}
\label{fig:baseline:matching}
\vspace{-0.5em}
\end{figure}

In practice, the similarity of a shape pair can be calculated by matching their skeleton graphs. For instance, the Path Similarity (PS) method~\cite{Bai2008PSS} aims to match skeleton endpoints using the similarity between their corresponding skeleton paths (Fig.~\ref{fig:baseline:matching} (left)). The final shape similarity is calculated by summing up the similarity of corresponded endpoints. Based on the idea of shape context~\cite{Belongie2002SMA}, the skeleton context (SC) method~\cite{Kamani2016SMU} employs log-polar histograms to describe sample skeleton points along the paths for matching (Fig.~\ref{fig:baseline:matching} (right)). The final shape similarity is computed by adding the distances between the matched skeleton points. We employ these methods to build baselines using our GTs on the Kimia216, MPEG400, Tetrapod120 and SwedishLeaves datasets (as they are actively used in this scenario). Specifically, we use each shape as a query and retrieve ten most similar shapes from the whole dataset according to their similarities. As shown in Table~\ref{tab:exp:baseline:retrieval}, the final value in each position (columns) is the total number of occurrences that matches query class at that position based on all shapes within a dataset. For example, the third position of the PS method on Kimia216 dataset shows that from 216 retrieved results in this position, 197 shapes have the same class as their query. We can see that the PS~\cite{Bai2008PSS} method achieves the best result among all the datasets, across all positions. This is because the PS method employs geodesic paths for matching endpoints, which makes it robust to scaling, rotation, occlusion, and also to same-class-objects of different topological structures.
\begin{table}[t!]
\centering
\setlength{\tabcolsep}{6.7pt}
\caption{Comparison of two skeleton-based shape retrieval methods using our GTs. PS: path similarity. SC: skeleton context.}
\label{tab:exp:baseline:retrieval}
\renewcommand{\arraystretch}{1.1}
\setlength{\arrayrulewidth}{1pt}
\begin{tabular}{ccccccccccc}
\hline
\textit{Kimia216}       & 1st & 2nd & 3rd & 4th & 5th & 6th & 7th & 8th & 9th & 10th \\ \hline
PS~\cite{Bai2008PSS}    & 216 & 206 & 197 & 185 & 173 & 169 & 154 & 150 & 140 & 119 \\
SC~\cite{Kamani2016SMU} & 196 & 91 & 80 & 82 & 77 & 71 & 70 & 72 & 61 & 57 \\
\hline
\textit{MPEG400}        & 1st & 2nd & 3rd & 4th & 5th & 6th & 7th & 8th & 9th & 10th \\ \hline
PS~\cite{Bai2008PSS}    & 400 & 384 & 380 & 367 & 363 & 351 & 339 & 338 & 327 & 318 \\
SC~\cite{Kamani2016SMU} & 382 & 131 & 148 & 164 & 162 & 150 & 144 & 142 & 117 & 123 \\
\hline
\textit{Tetrapod120}    & 1st & 2nd & 3rd & 4th & 5th & 6th & 7th & 8th & 9th & 10th \\ \hline
PS~\cite{Bai2008PSS}    & 120 & 110 & 90 & 86 & 73 & 73 & 72 & 55 & 55 & 51 \\
SC~\cite{Kamani2016SMU} & 113 & 48 & 33 & 24 & 34 & 32 & 18 & 31 & 23 & 21 \\
\hline
\textit{SwedishLeaves}  & 1st & 2nd & 3rd & 4th & 5th & 6th & 7th & 8th & 9th & 10th \\ \hline
PS~\cite{Bai2008PSS}    & 1125 & 974 & 914 & 881 & 868 & 833 & 813 & 790 & 785 & 767 \\
SC~\cite{Kamani2016SMU} & 1057 & 220 & 207 & 203 & 198 & 190 & 204 & 198 & 169 & 184 \\
\hline
\end{tabular}
\vspace{-0.5em}
\end{table}

For the MPEG7 and Animal2000 datasets, the Bulls-eye Scores (BES)~\cite{Latecki2000SDF} are normally computed for quantitative evaluation. BES is calculated as a ratio between the correctly matched shapes to the total number of possible matches. For instance, as there are 1,400 and 2,000 queries in MPEG7 (20 in each class) and Animal2000 (100 in each class) datasets, the total number of possible matches are $1400\times20$ and $2000\times100$, respectively. Accordingly, we employ GTs for skeleton-based shape retrieval. In addition to the PS and SC algorithms, the High-order (HO) matching method proposed in~\cite{Yang2020TAS} is also used in the evaluation. The HO method fuses similarities between the skeleton graphs with their geometrical relations characterized by multiple skeleton endpoints. Motivated by~\cite{Yang2020TAS,Bai2012COT,Kontschieder2010BPS}, experiments on both datasets are clustered into two groups: (1) pairwise matching similar to the experiments in Table~\ref{tab:exp:baseline:retrieval}, and (2) context-based matching by increasing the discrimination between different classes within the shape manifold. For this, the Mutual $k$NN Graph (MG)~\cite{Kontschieder2010BPS} and Co-Transduction (CT)~\cite{Bai2012COT} methods are employed (Table~\ref{tab:results:full:mpeg7}).

\begin{table}[t!]
\centering
\setlength{\tabcolsep}{3.9pt}
\caption{Bulls-eye scores (BES, \%) of three skeleton matching methods (PS: Path Similarity~\cite{Bai2008PSS}, SC: Skeleton Context~\cite{Kamani2016SMU} and HO: High-order Matching~\cite{Yang2020TAS}) based on MPEG7 (M7) and Animal2000 (A2) GTs. MG (Mutual $k$NN Graph~\cite{Kontschieder2010BPS}) and CT (Co-Transduction~\cite{Bai2012COT}) are their context-based extensions. The best scores are in boldface.}
\label{tab:results:full:mpeg7}
\renewcommand{\arraystretch}{1.1}
\setlength{\arrayrulewidth}{1pt}
\begin{tabular}{lccccccccc}
\hline
 & PS & PS+MG & PS+CT & SC & SC+MG & SC+CT & HO & HO+MG & HO+CT \\ \hline
M7 & 62.96 & 75.46 & 80.98 & 13.67 & 9.40 & 13.20 & 78.74 & 83.22 & \textbf{87.28} \\
A2 & 24.26 & 29.52 & 34.27 & 8.88 & 6.54 & 8.32 & 34.14 & 37.95 & \textbf{40.19} \\ \hline
\end{tabular}
\vspace{-0.5em}
\end{table}
For the pairwise experiments, we can clearly see that the HO method yields the best performance in both datasets. For the context experiments, we find that the BES improves after applying the MG and CT methods on the matching results from PS and HO. However, we find that both methods are ineffective on SC, with a decline in BES. The main reason is that similarity values between skeletons as calculated by the SC method are close to each other, and this results in poor shape retrieval performance: 13.67\% and 8.88\% on MPEG7 and Animal2000, respectively. Thus, the similarity values within skeletons of the same class are easily mixed with other classes.
\section{Discussion and Conclusion}
\label{s:con}
We present a brief overview of the challenges posed by the GT baselines and possible directions for future research.

\subsection{Analysis of Challenges}
\label{s:gt:challenges}
\begin{figure}[t!]
  \centering
    \includegraphics[width=0.68\linewidth]{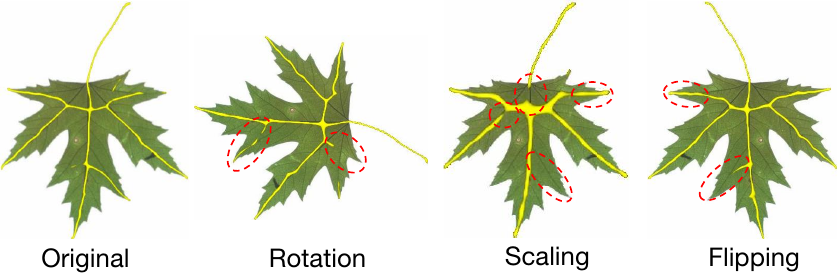}
  \caption{Evaluation of Ada-LSN~\cite{Liu2020ALS} on rotation, scaling and flipping.}
\label{fig:baseline:poor:leafs}
\vspace{-0.5em}
\end{figure}

\noindent{\bf Skeleton Extraction:} For most shape skeleton extraction approaches, we find that they cannot properly handle shapes with long and narrow (or lathy) regions. For instance, needle-like axopodia of actinophryid (EM200), petiole of leaves (SwedishLeaves), and antenna of insects (MPEG7). One possible solution is to generate their skeletons regionally, followed by integration and post-pruning steps. It is also interesting to note that most approaches are not guaranteed to preserve the topology of shapes containing holes. The simple way to resolve this issue is to fill the holes during the pre-processing step. Connected shapes with such kinds of complexity frequently occurs in the real-world but are rarely studied, {\it e.g.} class No. 10 in the SwedishLeaves dataset. Inspired by the theorems in~\cite{Bai2007SPB}, we suggest to incorporate boundary curves from both shapes and holes for skeleton extraction. For image skeleton extraction, the influence of the quality of training data is obvious. As a direct result of the consistent quality of our GTs, the F1 scores in Table~\ref{tab:exp:baseline:fmeasure} are generally higher than their original reported results~\cite{Xie2015HED,Ke2017SSR,Zhao2018HHF,Wang2019DFS,Liu2020ALS}. However, it is desirable for the community to introduce higher quality and larger scale datasets. Our GTs also capture richer dynamics that cannot be learned from existing datasets. For instance, we find that all CNN-based methods in Table~\ref{tab:exp:baseline:fmeasure} are sensitive to image rotation, scaling and flipping, which are fundamental requirements towards robust skeleton extraction in the real-world. However, there remains some inadequately addressed issues. As shown in Fig.~\ref{fig:baseline:poor:leafs}, even for Ada-LSN~\cite{Liu2020ALS} trained with data augmentation (rotation, scaling and flipping) on the SmithsonianLeaves dataset, we still find that some major skeletons branches are shortened, disconnected, and erased. For this, junction points, endpoints and skeleton graph could be encoded to restrict skeleton regression during the training period.
\begin{figure}[t!]
  \centering
    \includegraphics[width=0.8\linewidth]{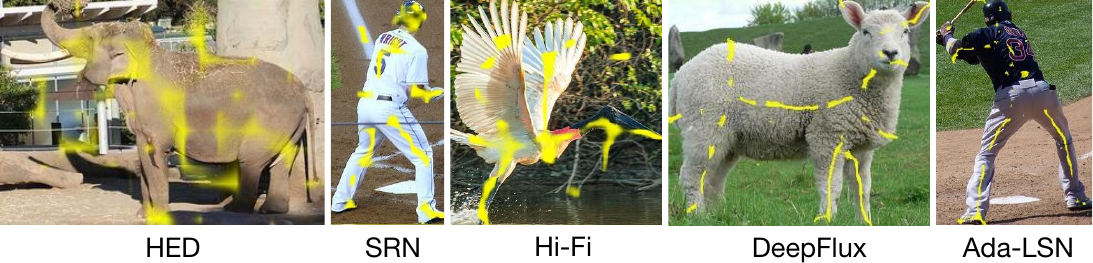}
  \caption{Poor detection samples from HED~\cite{Xie2015HED}, SRN~\cite{Ke2017SSR}, Hi-Fi~\cite{Zhao2018HHF}, DeepFlux~\cite{Wang2019DFS} and Ada-LSN~\cite{Liu2020ALS} methods.}
\label{fig:baseline:poor:samples}
\vspace{-0.5em}
\end{figure}

\noindent{\bf Skeleton Matching:} We further evaluate the pairwise matching algorithms in Table~\ref{tab:results:full:mpeg7} using the ArticulatedShapes dataset. This is because its GTs contain closed branches (Fig.~\ref{fig:gt:shapes} (b) (top)) with holes in tools such as scissors. We found that both the PS~\cite{Bai2008PSS} and HO~\cite{Yang2020TAS} algorithms cannot properly deal with skeleton graphs containing cycles. Though the SC~\cite{Kamani2016SMU} algorithm can be applied to such skeletons, it yields a poor performance with only 13.67 and 8.88 BES in MPEG7 and Animal2000, respectively. Therefore, we propose to improve the existing matching algorithms to support skeletons with closed branches. In Fig.~\ref{fig:baseline:poor:samples}, we see that most skeletons predicted in images are discontinuous with different widths and false positive points. In such cases, it is difficult to apply the existing algorithms for matching, classifying, and retrieval. In particular, these algorithms have been designed for one-pixel wide skeletons. To facilitate using image skeletons in practice, we propose to explore post-processing algorithms to bridge the gap between the image skeletons and the existing matching algorithms. Thus, a significant amount of research in future is necessary before image skeletons can become practically robust in many real-world objects.

\subsection{Conclusion}
We introduced a heuristic strategy for skeleton GT extraction in shape and image datasets. Our strategy is substantiated on both theoretical grounding and empirical investigation of human perception of skeleton complexity. To facilitate this, we developed a tool, SkeView, for skeleton GT extraction and used it on 17 existing image and shape datasets. We also systematically evaluated the existing skeleton extraction and matching algorithms to generate valid baselines using our GTs. Experiments demonstrate that our GT is consistent and can properly balance the trade-off between skeleton simplicity and completeness. We expect that the release of SkeView and the GTs to the community will benefit future research, particularly to address practical real-world challenges in CNN-based skeleton detectors and matching algorithms.

\bmhead{Acknowledgments}

Research activities leading to this work have been supported by the Natural Science Foundation of the Jiangsu Higher Education Institutions of China (Grant Number: 22KJB520008) and the Research Fund of Clobotics (Grant Number: KB1801ZW201609-03). We would like to thank Zixuan Chen from Darmstadt University of Technology (Germany) for his help in assembling the first version of SkeView.



\bibliography{cong}

\end{document}